\documentclass{isprs} 
\usepackage{setspace}
\usepackage{geometry} 
\usepackage{epstopdf}
\usepackage[labelsep=period]{caption}  
\usepackage[british]{babel} 
\usepackage[hang]{footmisc}
\usepackage{natbib}
\usepackage{booktabs} 
\usepackage{url} 
\usepackage{amsmath}
\usepackage{multirow}
\usepackage{subcaption}

\begin{document}

\title{Direct Estimation of Tree Volume and Aboveground Biomass Using Deep Regression with Synthetic Lidar Data}
\date{}


\author{
 Habib Pourdelan\textsuperscript{1}, Zhengkang Xiang\textsuperscript{1}, Hugh Stewart\textsuperscript{2}, Cam Nicholson\textsuperscript{3}, Martin Tomko\textsuperscript{1}, Kourosh Khoshelham\textsuperscript{1}
}

\address{
 \textsuperscript{1 }The University of Melbourne,(hpourdelan, zhengkangx)@student.unimelb.edu.au, (tomkom, k.khoshelham)@unimelb.edu.au\\
 \textsuperscript{2 }Private Forestry Service Victoria, Australia - htlstewart@gmail.com\\
 \textsuperscript{3 }Nicon Rural Services, Australia - cam@niconrural.com.au\\
}



\abstract{

Accurate estimation of forest biomass is crucial for monitoring carbon sequestration and informing climate change mitigation strategies. Existing methods often rely on allometric models, which estimate individual tree biomass by relating it to measurable biophysical parameters, e.g., trunk diameter and height. This indirect approach is limited in accuracy due to measurement uncertainties and the inherently approximate nature of allometric equations, which may not fully account for the variability in tree characteristics and forest conditions. This study proposes a direct approach that leverages synthetic point cloud data to train a deep regression network, which is then applied to real point clouds for plot-level wood volume and aboveground biomass (AGB) estimation. We created synthetic 3D forest plots with ground truth volume, which were then converted into point cloud data using a lidar simulator. These point clouds were subsequently used to train deep regression networks based on PointNet, PointNet++, DGCNN, and PointConv. When applied to synthetic data, the deep regression networks achieved mean absolute percentage error (MAPE) values ranging from 1.69\% to 8.11\%. The trained networks were then applied to real lidar data to estimate volume and AGB. When compared against field measurements, our direct approach showed discrepancies of 2\% to 20\%. In contrast, indirect approaches based on individual tree segmentation followed by allometric conversion, as well as FullCAM, exhibited substantially large underestimation, with discrepancies ranging from 27\% to 85\%. Our results highlight the potential of integrating synthetic data with deep learning for efficient and scalable forest carbon estimation at plot level.

}


\maketitle


\section{Introduction}

The increase in global temperatures, especially in the last 30 years, far exceeds what can be attributed to natural climate variations \citep{seinfeld2011insights}. Global heating has raised concerns about its consequences, such as rising sea levels, altered precipitation patterns, more frequent and intense extreme weather events, changes in agricultural yields, glacier retreats, species extinctions, expanded ranges of disease vectors, and other impacts \citep{change2007climate, upadhyay2020markers, singh2024global}. The rise in atmospheric carbon dioxide (CO$_2$) concentration is widely regarded as the primary factor driving global heating \citep{florides2009global}. In the 2015 Paris Agreement, 125 nations committed to limiting global warming to less than 2°C above pre-industrial levels \citep{unfccc2015report}. To achieve this goal given the current rates of temperature increase, immediate and substantial reductions in CO$_2$ emissions are necessary, targeting complete de-carbonization of the global economy by 2050 \citep{lal2008carbon}.

A mitigation strategy that is gaining significant attention in policy discussions is carbon offsets. This involves one entity reducing emissions or increasing greenhouse gas (GHG) sequestration to compensate for the emissions of another entity \citep{lovell2010governing}. Forest management, in particular, is being highlighted within carbon offsets as a method to reduce emissions or enhance the uptake and storage of CO$_2$. When integrated into a broader cap-and-trade program, forest offsets have the potential to offer low-cost GHG mitigation, thereby reducing the overall cost of implementing climate policies \citep{tavoni2007forestry}. It is widely agreed that forests are crucial in addressing climate change and working towards the goal of carbon neutrality \citep{wang2024factors}.

The forest ecosystem is the most expansive and vital natural habitat among terrestrial ecosystems \citep{becker2023country, hernandez2025improving}. It has a vital part in preserving the global ecological equilibrium, accounting for approximately 80\% of the Earth's plant biomass \citep{pan2013structure,li2020forest}. The world's forests contain about 45\% of terrestrial carbon \citep{bonan2008forests}. The majority of this carbon is stored in trees as aboveground biomass (AGB) through photosynthesis. The biomass of a forest refers to the total dry weight of all living and dead components of every tree per unit surface area \citep{vashum2012methods}. AGB, a part of the overall biomass, includes only plant parts that are above the ground, such as leaves, branches and stems \citep{ehlers2022mapping, santoro2024design}. Forest AGB significantly contributes to the total forest biomass, accounting for approximately 70\% to 90\% \citep{cairns1997root}. There is growing interest in the estimation of forest AGB since it is a crucial factor in assessing the forest’s ability to store carbon and maintain a balance of carbon \citep{soja2025sub, li2020forest, ni2025enhancing}. Approximately half of a tree's dry biomass corresponds to the amount of carbon stored within \citep{houghton2009importance}.
Estimating forests AGB is essential for quantifying the amount of carbon sequestered by trees. Accurate measurements of AGB enable the determination of the quantity of carbon stored in a given forest area. This information is crucial for creating reliable carbon credits, which represent the amount of CO$_2$ that has been sequestered from the atmosphere. In the carbon marketplace, carbon credits are traded as part of emission trading systems or voluntary carbon offset programs. These credits are generated based on the amount of carbon sequestered through various activities, including afforestation, reforestation, and forest conservation. By accurately estimating forest AGB, we can ensure that the carbon credits generated are a true representation of the carbon stored in forests \citep{mardiatmoko2018forest, fahey2010forest, reyer2009climate}.

Various methods have been used for AGB estimation, each with its advantages and limitations. Remote sensing methods, including optical imagery, RADAR (RAdio Detection And Ranging), and lidar (light detection and ranging), potentially offer more efficient alternatives than ground-based methods \citep{zhu2015improving}. Among these, lidar is particularly notable for its accuracy and ability to capture the 3D canopy structure. By emitting laser pulses and measuring their return time, lidar generates detailed 3D models of forest structure, capturing both canopy and ground-level features.  
This enhances the precision and accuracy AGB estimation, particularly across extensive and densely forested regions \citep{tian2023review, khan2024forest, rana2014effect}.
Compared to other remote sensing methods, lidar provides the highest resolution and accuracy, making it the preferred choice for AGB estimation \citep{popescu2007estimating}. 
Existing methods for AGB estimation from lidar data predominantly extract biophysical parameters of individual trees, such as diameter and height, and use allometric equations to convert these to AGB. 
Recent deep-learning \citep{lecun2015deep} instance-segmentation methods can now segment individual trees from lidar point clouds automatically, and some models generalize better across different sensors and survey platforms. These include sensor- and platform-agnostic approaches such as SegmentAnyTree \citep{wielgosz2024segmentanytree}, end-to-end transformer-based frameworks such as ForestFormer3D \citep{xiang2025forestformer3d}, and fully automated pipelines trained on labelled tree instances such as TreeLearn \citep{henrich2024treelearn}. These methods can yield tree-level segments that support downstream measurement of tree height at scale, enabling plot AGB estimates by summing per-tree allometric predictions rather than relying solely on canopy-height metrics or stand-level regressions.
Yet, even with strong segmentation models, this remains an indirect estimation chain in which errors accumulate: instance segmentation can still suffer from under- and over-segmentation in structurally complex stands with overlapping crowns, multi-layer canopies, variable point density, and occlusion, and these effects are well documented as persistent failure modes in current methods \citep{Luo2021individual}. The second source of uncertainty is the AGB conversion itself, because allometric equations are empirical approximations fitted to limited destructive samples, and their validity can shift with species group, site conditions, tree form, and measurement noise in the predictors \citep{park2026combining}. Consequently, high-quality tree segmentation is necessary but not sufficient for reliable AGB estimation and the approach still depends on the assumptions embedded in the allometry and on how robustly the segmentation outputs translate into unbiased height distributions across canopy layers and forest types.
With the remarkable success of deep learning in various recognition and estimation tasks, the application of deep learning models to tree AGB estimation has attracted significant attention in recent years. However, deep learning models require large volumes of data for training and are constrained by the lack of reliable ground truth data. Ground truthing involves validating AGB estimates with on-site measurements, which is essential for calibrating and verifying the accuracy of deep learning models. The absence of comprehensive and consistent ground truth data can lead to uncertainties in AGB estimates, limiting the overall reliability and applicability of lidar-derived AGB for large-scale forest management and carbon stock assessments \citep{borsah2023lidar}.

\begin{figure}
\centering
\includegraphics[width=3.7in]{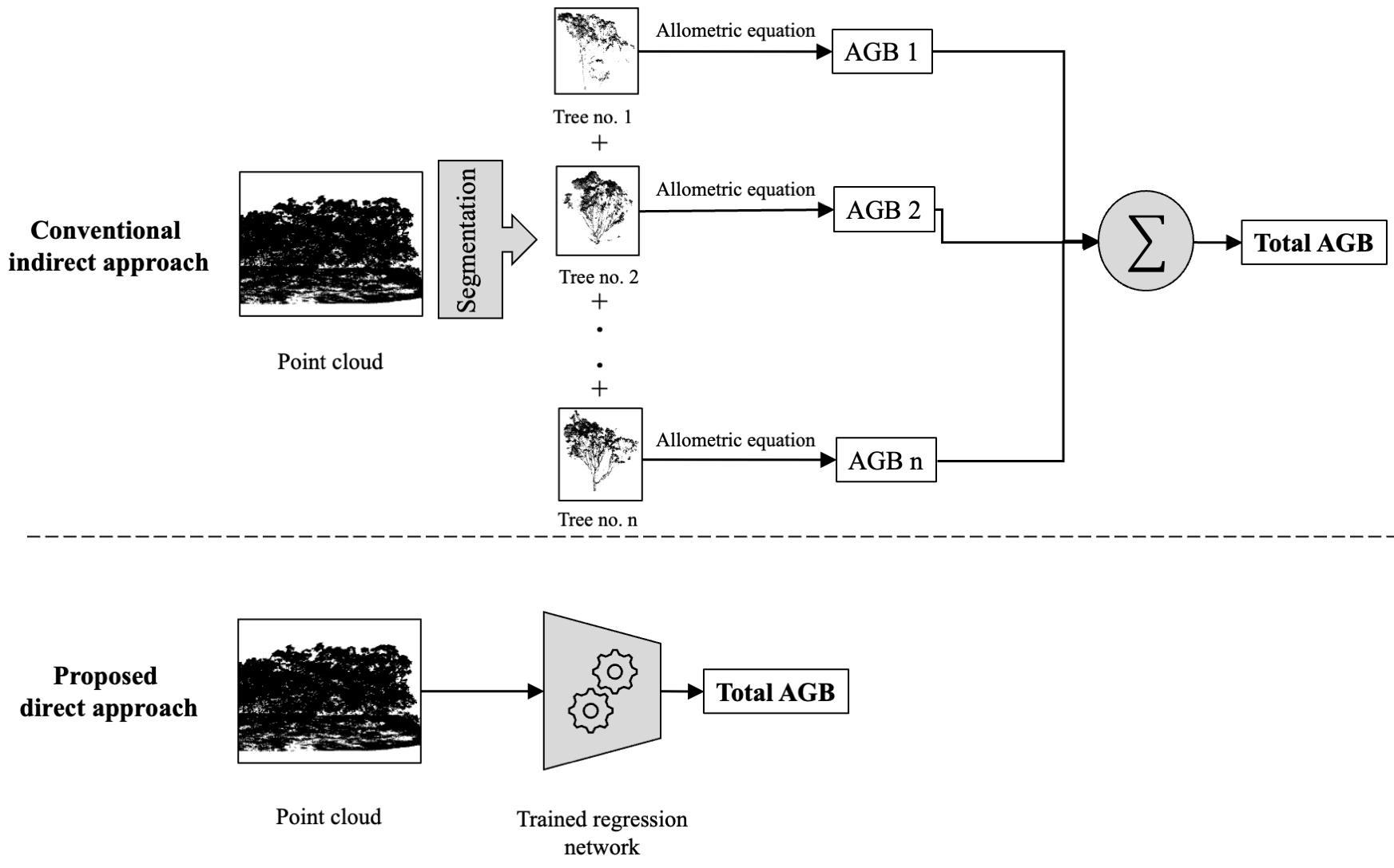}
\caption{Schematic comparison of conventional indirect AGB estimation (segmentation + allometry + aggregation) versus the proposed direct plot-level regression approach.}
\label{fig:directvsindirect}
\end{figure}

To address these challenges, we propose a direct method for plot-level tree volume and AGB estimation from LiDAR point clouds that does not require individual-tree segmentation. In contrast to conventional indirect pipelines—where trees are first delineated and per-tree attributes are converted to AGB via allometric equations and then aggregated (Figure~\ref{fig:directvsindirect}), our approach regresses plot-level volume/AGB end-to-end from the point cloud.
To overcome the lack of accurate AGB ground truth, we develop a deep regression network trained on synthetic data. The proposed method involves creating synthetic forest plots with exact ground truth volume and scanning these with a simulated lidar sensor to generate synthetic point clouds. The trained models are then applied to real lidar surveys by subdividing each site into tiles analogous to the synthetic plots, predicting wood volume per tile, and aggregating tile outputs to obtain site-level wood volume, AGB, and carbon estimates. This approach not only enhances the reliability of AGB estimates but also leverages deep learning to improve the accuracy and efficiency of lidar-based forest carbon estimation.

The contributions of the present study include:

\begin{itemize}
    \item A machine learning approach for accurate estimation of tree volume and AGB directly from point clouds. To the best of our knowledge, this is the first method for AGB estimation that does not rely on point cloud segmentation and allometric models.
    \item A framework for generating synthetic data with exact ground truth for training a deep regression network.
    \item A comparative evaluation of four deep learning models conducted to assess the feasibility and robustness of the proposed approach, and demonstrate that deep regression networks trained on synthetic data can provide accurate estimates of AGB and carbon stock for real trees at plot level; and
    \item An investigation of effective downsampling strategy for point cloud analysis, comparing the farthest point sampling with random sampling strategies.

\end{itemize}

The paper is structured as follows: Section 2 reviews related literature, Section 3 explains materials and methods (including synthetic lidar data, preprocessing, and deep learning models for AGB estimation), Section 4 presents results, Section 5 discusses findings, comparisons, and implications, and Section 6 outlines conclusions, limitations, and future research directions.

\section{Literature Review}\label{sec:litrature}

In this section, we provide comprehensive information about forest AGB estimation, focusing particularly on the advancements and applications of lidar technology. We review various methods of AGB estimation, from traditional field-based techniques to modern remote sensing approaches. Special attention is given to the integration of deep learning models with lidar data, highlighting recent innovations and their impact on the accuracy and efficiency of AGB estimation.
The most accurate way to acquire forest AGB data is through destructive sampling, which entails collecting, drying, and weighing live and dead AGB. However, the destructive method is expensive, challenging in regions with limited accessibility, and difficult for obtaining comprehensive spatial and temporal dynamics of AGB information \citep{stovall2017non}. Non-destructive methods for estimating forest AGB can be classified into three categories: field measurements, ecological model simulations, and remote sensing-based approaches. Field measurements involve the development of allometric equations utilizing tree height and diameter data obtained from forest inventory data or supplementary field plots. However, conducting these measurements at a regional scale poses significant challenges due to the time-consuming and labor-intensive process of ground-based data collection \citep{ketterings2001reducing, seidel2011review}. Ecological model simulation helps researchers to study ecosystems, but some challenges need attention. This method is frequently limited to specific locations and demands numerous input parameters, the accurate values for which may be challenging to acquire \citep{tian2023review}. Because of the limitations associated with the aforementioned methods for regional AGB estimation, remote sensing has been extensively employed in recent decades due to its capacity for broad-area coverage. Consequently, remotely sensed data continue to be the primary data sources for mapping AGB \citep{ali2023assessment, gibbs2007monitoring}.

Various sensors and methods utilized in remote sensing provide means for estimating AGB at a reasonable cost and with acceptable accuracy. This approach combines ground measurement data with data obtained from remote sensors to estimate parameters closely associated with AGB. Passive optical imagery, synthetic aperture RADAR (SAR), and lidar are three remote sensing methods utilized for gathering data for AGB estimation \citep{ali2023assessment}. Optical remote sensing signals have limited penetration capabilities, primarily capturing data on the horizontal layout of vegetation, consequently providing inadequate representation of its vertical structure, Also, it heavily relies on the absence of clouds \citep{sinha2015review}. 
SAR instruments are capable of continuous forest monitoring and deep penetration, as their longer wavelength electromagnetic energy is not affected by cloud cover, unlike optical sensors.  Various studies have shown significant correlations between SAR backscattering coefficients and forest AGB \citep{stelmaszczuk2018estimation,peregon2013use}. However, SAR data are influenced by topographical variances (such as steep slopes or cliffs). Also, high-frequency SAR data can cause saturation problems, particularly in areas with high forest AGB \citep{sinha2015review, zeng2022forest}. 

Lidar is a method for acquiring precise three-dimensional data of terrain surface including objects (such as trees and buildings) and topography. The laser pulses emitted by a lidar sensor can penetrate forest canopies, delivering detailed information on vertical structures. This data can be used to extract critical parameters such as tree height, canopy volume, and crown dimensions, which are directly related to AGB. The comprehensive vertical profile provided by lidar allows for precise measurement of tree structures without the saturation issues commonly associated with other remote sensing methods, such as optical or radar imagery. Consequently, lidar-derived metrics have shown strong correlations with field-measured AGB, making it a preferred method for AGB estimation across various forest types and conditions \citep{zolkos2013meta, nikcombination, urbazaev2018estimation}.\\

Several methods have been employed to estimate AGB using lidar data, but most still depend on indirect modeling chains that constrain achievable accuracy. One common approach involves generating statistical variables or metrics from the canopy height model (CHM) or laser returns. This process yields a variety of lidar metrics such as dominant height, mean height, 3D point cloud density at various height percentiles, and canopy cover. Additional metrics might include canopy volume, skewness, kurtosis, and other descriptive statistics that capture the vertical and horizontal structure of the forest \citep{borsah2023lidar}. To enhance the accuracy of AGB estimation, these lidar-derived metrics are often combined with forest attributes obtained from inventory plots, such as tree species, diameter at breast height (DBH), and tree age. This integration provides a more comprehensive dataset that reflects both the structural and compositional aspects of the forest. These combined datasets serve as inputs for various prediction models designed to estimate AGB. These models include linear and non-linear regression, random forests, support vector machines, and neural networks \citep{dhanda2017optimizing, dantas2021machine, kc2024estimation, li2022research, cao2018integrating}. 
However, this indirect approach results in the loss of essential structural information embedded within three-dimensional data. 

More recently, the community has increasingly moved toward individual tree segmentation as a route to more detailed AGB accounting. In principle, these outputs enable plot-level AGB estimation by summing tree-level AGB values derived from each segmented instance via allometric equations \citep{wielgosz2024segmentanytree, luo2025estimation, jung2025individual}. While individual tree segmentation has seen significant technical maturation, its utility for AGB estimation remains heavily contingent on the data acquisition mode and the inherent limitations of the downstream modeling chain. In Terrestrial Laser Scanning (TLS) point clouds, the highly detailed and precise 3D point clouds allow for the detailed 3D reconstruction of tree architecture, enabling researchers to calculate wood volume directly through quantitative structure models (QSMs) \citep{hackenberg2015simpletree, raumonen2013fast, demol2021forest, kunz2017comparison}. However, for aerial platforms such as Unmanned Aerial Vehicle (UAV) laser scanning (ULS) or Airborne Laser Scanning (ALS), the top to down capture is frequently limited by canopy occlusion, making full structural reconstruction of every branch virtually impossible \citep{kankare2013single, white2016remote}. Consequently, these platforms must rely on extracting biophysical features, such as tree height and crown diameter—to serve as inputs for empirical models. Even when utilizing state-of-the-art deep-learning \citep{henrich2024treelearn, xiang2025forestformer3d, wielgosz2024segmentanytree} frameworks can identify individual instances with high precision even in complex, closed-canopy environments, the final estimate is still limited to the allometric bottleneck. Because allometric equations are inherently approximate scaling laws derived from limited destructive samples, they introduce systematic errors that a geometric segmenter cannot rectify. This indirect estimation chain fails to leverage the full 3D point distribution, ensuring that even "perfect" segmentation remains hostage to the assumptions and measurement noise embedded in traditional allometry \citep{park2026combining}.

In contrast, deep neural networks with regression layers trained on lidar-derived point clouds 
offer improved accuracy and efficiency in AGB estimation while maintaining the detailed structural integrity of the original data. \citet{oehmcke2024deep} applied advanced deep learning techniques to estimate AGB, wood volume, and carbon stocks directly from airborne lidar point clouds. They adapted three neural network architectures, such as PointNet, KPConv, and Minkowski CNNs, to perfrom classification and regression tasks. Using a dataset that combined field measurements with lidar data and allometric models for ground truth, they found that these deep learning models significantly outperformed existing methods in estimating wood volume, AGB, and carbon stocks. 
\citet{seely2023modelling} compared two point cloud-based deep neural networks (DNNs), the Dynamic Graph Convolutional Neural Network (DGCNN) and the Octree-based Convolutional Neural Network (OCNN), with a random forest model for direct AGB estimation. The DNNs demonstrated marginally superior performance compared to the random forest model, with OCNN accounting for 5\% more variance in the data (R² = 0.76) and reducing the Mean Absolute Percentage Error (MAPE) by 20\%.

However, these methods require a large dataset of lidar data and corresponding field data to learn the relationship between point cloud metrics and forest characteristics. Gathering field data is both labor-intensive and expensive, resulting in a limited number of field samples collected in smaller areas as studied by \citet{schafer2023generating}.
For smaller areas, \citet{laino20243dfin} introduced 3DFin, a software tool designed for automated 3D forest inventories using TLS, providing capabilities for accurate tree detection, DBH estimation, and height measurement. While 3DFin significantly streamlines the traditionally labor-intensive forest inventory tasks, its reliance on TLS data makes it unsuitable for vast forest landscapes. 
\citet{laurin2014above} indicate that high-quality ground truth data, particularly with precise geo-location and larger plot sizes, is essential for planning lidar-based AGB estimates. It is even more crucial when airborne lidar is employed as an intermediate step to upscale field-measured AGB to a larger area or region, as this procedure introduces additional uncertainties. Several studies have used synthetic lidar data to overcome the limitations of acquiring ground truth data for AGB estimation. \citet{schafer2023generating} developed a method to create synthetic ALS point clouds using the HELIOS++ simulation framework by first generating forest scenes with tree positions, species, and sizes, and then scanning opaque voxelized point clouds. \citet{song20233d} created 3D tree models using Arbaro and simulated lidar point clouds for various forest scenes, applying machine learning models to estimate AGB with improved accuracy through data augmentation. Similarly, \citet{wang2020unsupervised} used SpeedTree software to create synthetic forest plots and applied HELIOS++ to convert them into point clouds for single-tree isolation and classification, achieving high accuracy in tree structure measurements that can be used for AGB estimation. These approaches provide a controlled, scalable way to generate reliable lidar data for AGB studies without the need for extensive ground truth data.
Utilizing synthetic lidar data provides reliable ground truth for training and validating deep learning models in situations where acquiring extensive and diverse real point cloud data is challenging or cost-prohibitive. The creation and use of synthetic datasets not only fill the gaps in available real data but also enable the simulation of diverse scenarios that may not be easily accessible in the real world, thus broadening the scope of training and enhancing model robustness. \citet{feng2025spread} introduced SPREAD, a large-scale, photo-realistic synthetic dataset built with Unreal Engine 5, providing multimodal data (RGB, depth, point clouds, segmentation labels, and tree parameters) that significantly improves trunk and canopy segmentation performance through pretraining and hybrid training, while reducing reliance on expensive real-world datasets. \citet{bryson2023using} developed a tree simulation framework to generate synthetic lidar point clouds, allowing PointNet++ models to out-perform, or provide comparable performance to, models trained on limited real datasets, particularly for stem segmentation across diverse forest sites. \citet{xiang2024synthetic} leveraged deep generative models to create synthetic lidar point clouds, which substantially improved object recognition performance by augmenting the limited variability and volume of real training datasets. Similarly, in indoor settings where GPS is unreliable, \citet{zhao2024moli} developed MoLi-PoseNet which utilized synthetic lidar scans with accurate lidar pose to predict the pose of real lidar scans. These examples illustrate the broad applicability and significant benefits of synthetic data across various domains. Likewise, \citet{chitnis2021generating} adopted this approach by employing a 3D Adversarial Autoencoder (3dAAE) to generate synthetic point segments, addressing the common challenges associated with sparse and inadequately labeled mobile lidar data. Their method significantly enhanced the classification accuracy of urban scene elements such as vehicles, pedestrians, and traffic signs. This improvement underscores the value of synthetic data in overcoming the inherent limitations of sparse datasets, further validating the use of synthetic lidar data as a substantial resource for training deep learning models more comprehensively. Thus, synthetic lidar data proves to be an invaluable asset in training and validating models where real data is scarce or incomplete, facilitating advancements in both object recognition in autonomous vehicles and sensor localization in indoor navigation.

\begin{figure}
\centering
\includegraphics[width=3.3in]{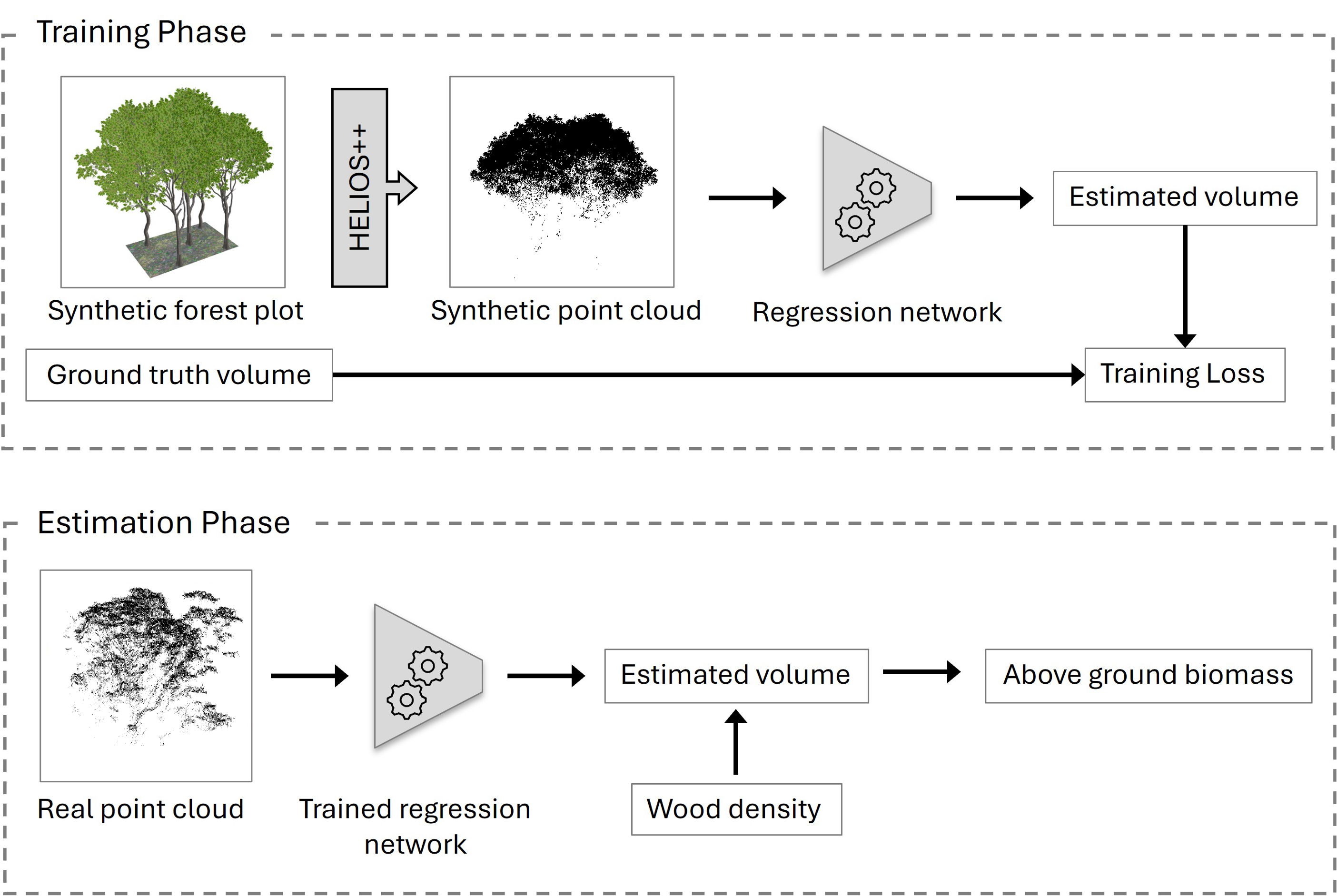}
\caption{Overview of the proposed methodolog}
\label{fig:overview}
\end{figure}

\section{Materials and Methods}\label{sec:methodology}
We propose a method for direct estimation of tree volume and AGB from lidar data using a deep regression network trained on synthetic lidar point clouds. The process involves generating realistic 3D synthetic forest plots and converting these into synthetic lidar data that accurately simulate real-world forest environments. Deep learning models are then trained on this synthetic point cloud data to estimate wood volume. Once validated, the models are applied to real-world lidar data. The estimated wood volume is subsequently converted to AGB and carbon stocks using species-specific wood density values \citep{brede2019non}. Figure \ref{fig:overview} provides an overview of the proposed method.
In the following sections, we detail the components of the proposed method. Section 3.1 introduces the study area and the lidar data used for AGB estimation. Section 3.2 describes the generation of synthetic lidar data, which simulates realistic forest environments to provide a robust dataset for model training and validation. Section 3.3 presents the deep learning models used for volume estimation, including their architecture and the specific adjustments made to adapt them for point cloud regression tasks. Finally, Section 3.4 outlines the training and evaluation parameters used for the deep learning models. Finally, Section 3.5 describes the indirect (allometry-based) baselines, including individual tree segmentation baselines and the Full Carbon Accounting Model (FullCAM) \citep{FullCAM_Reference}, used for comparative evaluation. 

\subsection{Study area}
\begin{figure}
\centering
\includegraphics[width=3.3in]{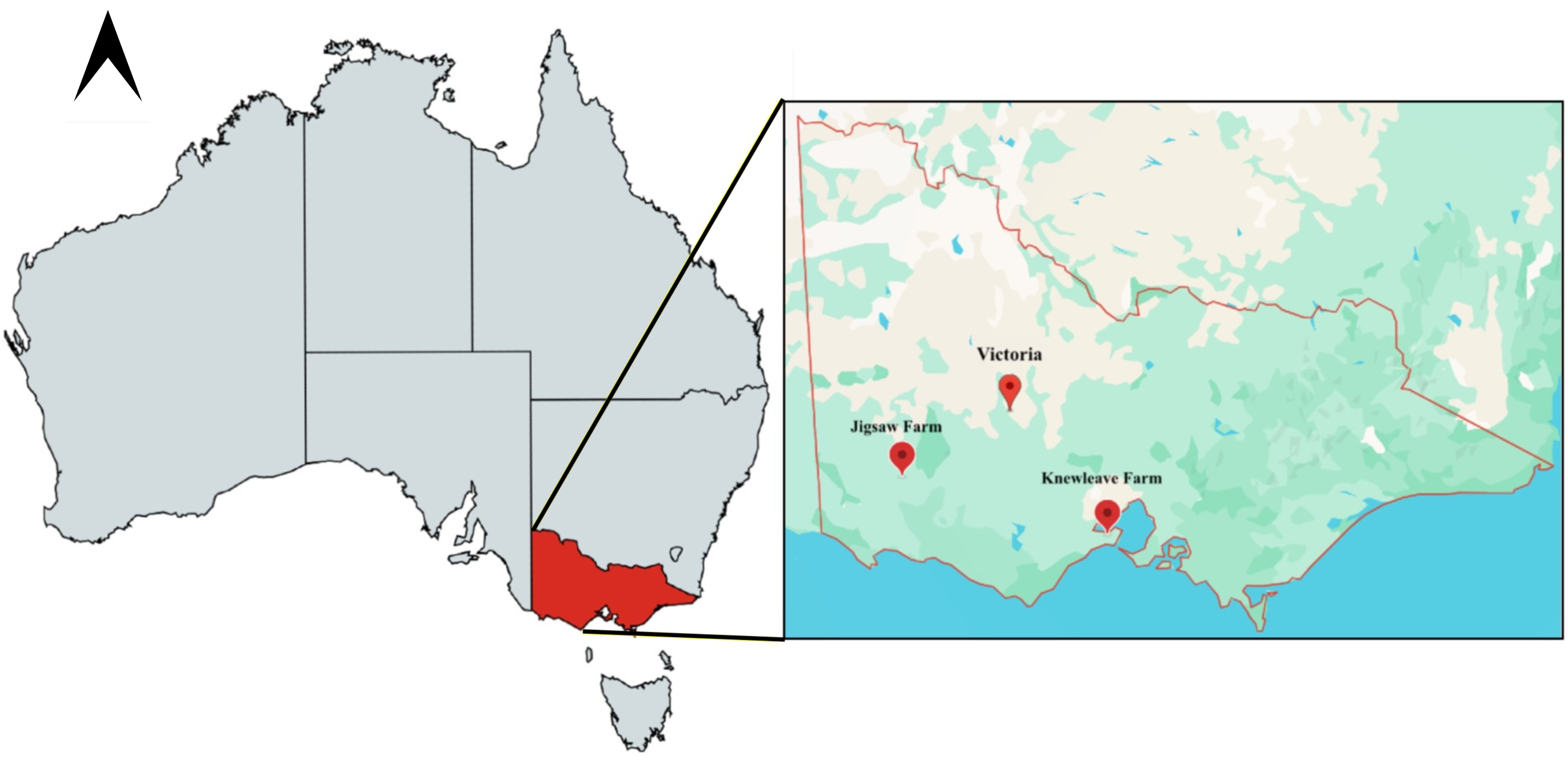}
\caption{Lidar data and field measurement of AGB were collected in two study areas, Knewleave and Jigsaw Farms, both located in Victoria, Australia.}
\label{fig:area}
\end{figure}

The study was conducted at two farms in Victoria, Australia, as shown in Figure \ref{fig:area}: Jigsaw Farms in the southwest, where we focused on Melville Forest and also sampled a site at Hensley Park, and Knewleave Farm on the Bellarine Peninsula in the south. Both farms are sheep and cattle grazing enterprises. A feature of both properties is extensive tree planting integrated into daily agriculture to provide shade, shelter, biodiversity, and timber products. At Jigsaw Farms, we sampled three sites at Melville Forest and one at Hensley Park, with areas ranging from 0.8 to 1.8 hectares, while at Knewleave farm, a single site covering 2.5 hectares was sampled. These plantings featured a mix of seedling and direct-seeding methods and represented closed forest conditions with at least 80\% canopy projection based on spatial imagery.
The sampling followed the methodology outlined by \citet{bennett2022assessing}, using three randomly located belt transects with a width of 5 meters and length of 30 meters. Within each plot, the stem diameter of all standing trees with diameter greater than 5 cm was measured, and the species group of each tree was recorded. 
Following the completion of field data collection, the AGB for each sampled site was estimated using established allometric equations based on stem diameter measurements. These allometric equations were used only to derive the field-based reference (ground-truth) AGB values for validation and were not used as inputs to the proposed direct regression models. We utilized the following allometric equation for Eucalyptus species as outlined by \citet{paul2016testing}. 

\begin{equation}
    AGB = \exp(ln(a) + b \cdot \ln(D)) \cdot c
\end{equation}
where AGB represents the estimated AGB (kg) and D is the measured stem DBH (cm). Other parameters are provided at Table \ref{tab:allometric_coefficients}. All trees measured were within the respective domains of the allometric models.

\begin{table*}[t]
\caption{Coefficients for allometric models for AGB using a predictor of diameter (D) measured at either 10 cm or 130 cm height.}
\label{tab:allometric_coefficients}
\centering
\begin{tabular*}{\textwidth}{@{\extracolsep{\fill}} lllcc @{}}
\toprule
\textbf{Model (domain in} & \textbf{Diameter} & \boldmath $\ln(a)$ & \boldmath $b$ & \textbf{Correction} \\
\textbf{parentheses)} & \textbf{measurement} & & & \textbf{factor (c)} \\
\midrule
Eucalypt ($D_{130} < 169$ cm) & $D_{130}$ & -2.016 & 2.375 & 1.0668 \\
Multi ($D_{10} < 62$ cm)       & $D_{10}$  & -2.757 & 2.474 & 1.0775 \\
Shrub ($D_{10} < 50$ cm)       & $D_{10}$  & -3.007 & 2.428 & 1.1281 \\
Other ($D_{130} < 102$ cm)     & $D_{130}$ & -1.693 & 2.220 & 1.0436 \\
\bottomrule
\end{tabular*}
\end{table*}

To facilitate carbon stock analysis, these AGB estimates were standardized to tonnes per hectare. These values were subsequently converted to the mass of carbon in the AGB using a standard multiplication factor of 0.5. This field-derived carbon sequestered value served as the primary ground truth for evaluating the performance of our direct deep regression models and the state-of-the-art segmentation methods.

\subsection{Synthetic data generation}
The synthetic data generation process involves several key steps, elaborated as follows and illustrated in Figure \ref{fig:synthetic_data}, which outlines the workflow from generating synthetic forest plots to converting them into point clouds for model training.
\subsubsection{Creating 3D synthetic forest plot}
\begin{figure*}[t] 
\centering
  \begin{subfigure}[b]{0.24\linewidth}
    \centering
    \includegraphics[width=\textwidth]{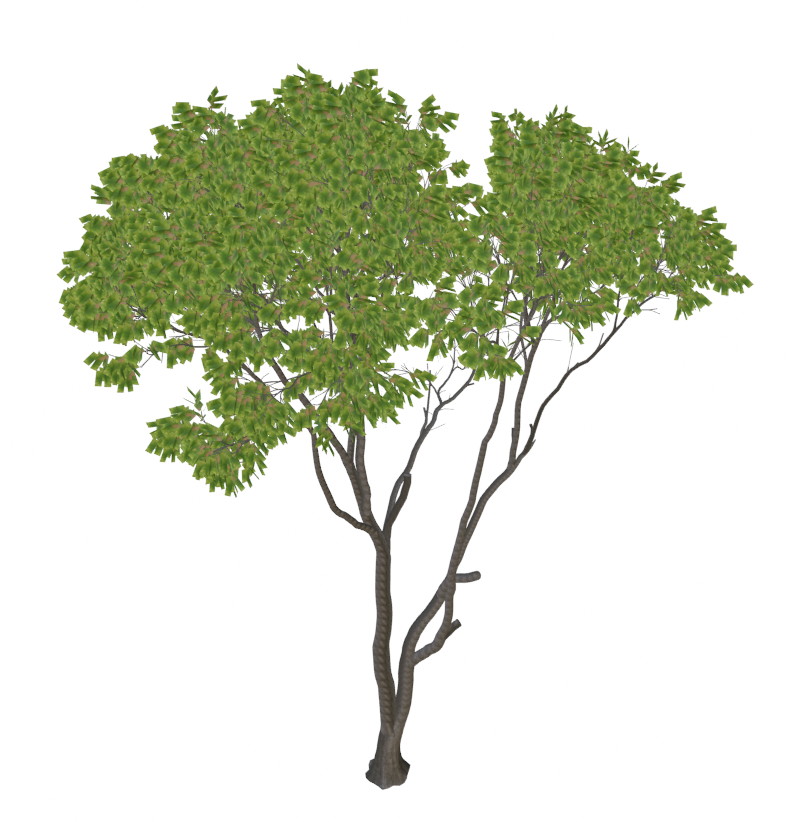}
    \caption{}
    \label{fig:tree}
  \end{subfigure}
  \hfill
  \begin{subfigure}[b]{0.26\linewidth}
    \centering
    \includegraphics[width=\textwidth]{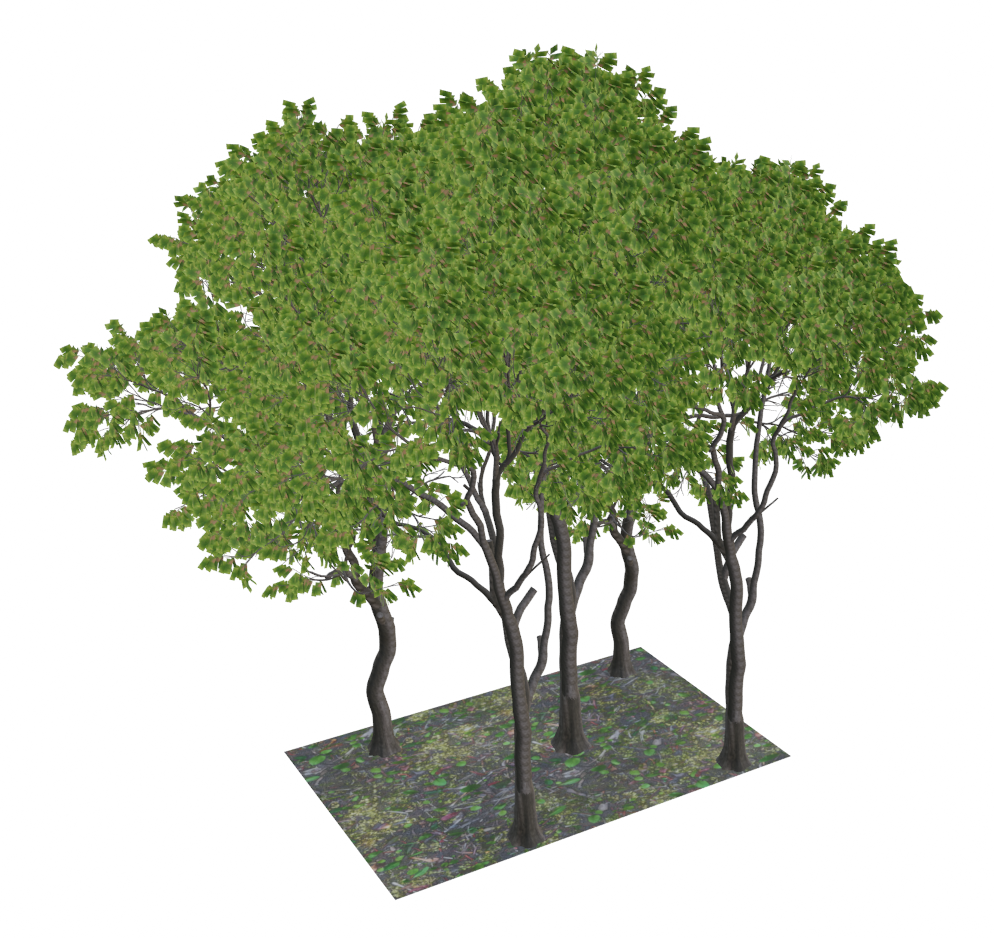}
    \caption{}
    \label{fig:plot}
  \end{subfigure}
  \hfill
  \begin{subfigure}[b]{0.3\linewidth}
    \centering
    \includegraphics[width=\textwidth]{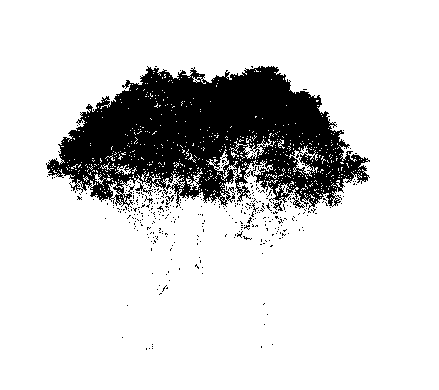}
    \caption{}
    \label{fig:pointcloud}
  \end{subfigure}
\caption{Synthetic data generation process. (a) 3D eucalyptus tree model, (b) Synthetic forest plot built in Blender, (c) Simulated point cloud.}
\label{fig:synthetic_data}
\end{figure*}

Since eucalyptus trees were the dominant species at our selected site, seven distinct 3D eucalyptus tree models \citep{yankobe2023eucalypt} were used to generate synthetic forest plots, each differing in height, trunk diameter, number of leaves, and branches. Figure \ref{fig:synthetic_data}-a illustrates one of the generated tree models used in the plot construction. Utilizing Blender version 3.6.2 \citep{blender}, a total of 1200 plots were generated. Each plot contains a variable number of eucalyptus trees, each exhibiting unique characteristics. As shown in Figure \ref{fig:synthetic_data}-b, the plots have a rectangular shape, and the dimensions of each plot were adjusted based on the number of trees. During plot creation, factors such as tree density, average tree height at the actual site, and the specific characteristics of the eucalyptus species were considered.

\subsubsection{Volume calculation of each plot}
To determine the wood volume of each plot, we first calculated the wood volume of each tree. First, each part of the leaf-off tree such as the trunk and branches was segmented into separate components using Blender software. Using the Python libraries Trimesh developed by \citet{trimesh} and PyMeshFix developed by \citet{attene2010lightweight}, the volume of each component was calculated. Trimesh facilitates the handling, transformation, and visualization of triangular meshes, while PyMeshFix ensures that these meshes are watertight and manifold, which is essential for accurate volume estimation. The total volume of each tree was determined by summing the volumes of its individual parts. Subsequently, the total wood volume for each plot was obtained by summing the volumes of all trees within the plot.
\subsubsection{Converting 3D plots to point cloud using laser scanning simulation}
Each 3D forest plot is converted into a point cloud using HELIOS++ v1.3 \citep{esmoris2022virtual}. HELIOS++ is a general-purpose software package developed in C++11 for simulating terrestrial, mobile, and airborne laser scanning surveys.
Since our real data was collected using ULS, we scanned the synthetic forest plots using ULS in our simulation to ensure consistency and comparability between the synthetic and real point cloud datasets. We selected the Riegl VUX-1UAV22 as the simulated scanner due to limitations in the Helios++ scanner library, which, at the time, did not support 2D multi-beam scanners, specifically the Hovermap ST, which we had used for collecting real data. The settings of Riegl VUX-1UAV22 were adjusted to achieve a point density approximately equal to our real lidar data. Table 1 presents the acquisition settings of the laser scanners used for collecting real point cloud data and generating synthetic datasets. A key distinction between Hovermap ST and the Riegl VUX-1UAV22 scanner lies in their scanning methodologies. The Riegl VUX-1UAV22 is a 2D aerial laser scanner that primarily scans within a single plane, whereas Hovermap ST uses a 2D multi-beam approach,  where laser beams are emitted within multiple approximately parallel planes. This multi-plane capability enhances Hovermap ST's capacity to gather volumetric information, making it highly versatile for applications that require detailed spatial data.  Despite these differences, the Riegl VUX-1UAV22 provided a suitable alternative with robust 3D scanning capabilities, effectively meeting the data needs of the study.  
Figure \ref{fig:synthetic_data}-c shows an example point cloud generated using HELIOS++.

\begin{table*}[t]
\caption{Acquisition settings of the real (HOVERMAP ST) and simulated (Riegl VUX-1UAV22) laser scanners}
\label{tab:acquisition_settings}
\centering
\begin{tabular*}{\textwidth}{@{\extracolsep{\fill}} l c c }
\toprule
\textbf{Setting} & \textbf{Riegl VUX-1UAV22} & \textbf{HOVERMAP ST} \\
\midrule
Altitude (m) & 35 & 30--40 \\
Flight speed (m/s) & 2 & Not available \\
Pulse repetition frequency (kHz) & 200 & 300 \\
Scan frequency (Hz) & 80 & 20 \\
Max. returns per pulse & 5 & 2 \\
Scan angle (°) & 180 & 360 \\
\bottomrule
\end{tabular*}
\end{table*}
\subsubsection{Data preparation}
We initially generated a dataset consisting of 1200 plots. To ensure consistency across all data samples and facilitate efficient batch processing, the point cloud data for each plot was uniformly downsampled to 2048 points. This fixed input size is particularly important for deep learning models designed for point cloud data, as they typically require a standardized number of points to process data effectively. Two distinct downsampling methods were employed to evaluate their impact on model performance \citep{li2023comparative}.
The first method employed the FixedPoints function from the torch-geometric library for random sampling, as developed by Fey and Lenssen \cite{fey2019fast}.
This function is designed to uniformly downsample point clouds to a fixed number of points. It achieves this by randomly sampling a specified number of points from the input point cloud without replacement, ensuring that the downsampled point set is a subset of the original data. However, this method does not consider the spatial distribution or geometry of the points when selecting the subset and it simply picks points at random.
The second method utilized farthest point sampling, ensuring a more spatially representative selection of points within each plot.
farthest point sampling operates iteratively by first randomly selecting an initial point and then repeatedly selecting the point that is farthest away (in Euclidean space) from all previously selected points. This process continues until the desired number of points is reached. By prioritizing spatial diversity, farthest point sampling ensures that the sampled points are evenly distributed across the input point cloud, effectively preserving its geometric structure and important features \citep{liu2022tree}.

To compare the spatial characteristics of downsampled point clouds produced by random sampling and farthest point sampling, we computed plot area, plot volume, point density per area, point density per volume, and average point spacing for each plot. Let $\{(x_i,y_i,z_i)\}_{i=1}^{N}$ denote the $N=2048$ points in a plot after downsampling, and let $x_{\min},x_{\max}$ (similarly for $y$ and $z$) denote the coordinate bounds.

The plot area is defined by the axis-aligned bounding box in the horizontal plane:
\begin{equation}
A = (x_{\max}-x_{\min})(y_{\max}-y_{\min}).
\label{eq:area}
\end{equation}

The plot volume is defined as the area times the vertical extent:
\begin{equation}
V = A (z_{\max}-z_{\min}).
\label{eq:volume}
\end{equation}

Point density per area and per volume are computed as:
\begin{equation}
\rho_{A} = \frac{N}{A}.
\label{eq:density_area}
\end{equation}

\begin{equation}
\rho_{V} = \frac{N}{V}.
\label{eq:density_volume}
\end{equation}

To quantify local sampling resolution, we computed average point spacing as the mean nearest-neighbour distance across all points. Nearest-neighbour distances were obtained using Open3D's \texttt{compute\_nearest\_neighbour\_distance}, which constructs a KD-tree and performs a $k$-nearest neighbour search with $k=2$ (ignoring the self-match) \citep{zhou2018open3d}. Let $d_i$ denote the Euclidean distance from point $i$ to its closest distinct neighbour; the average spacing is:
\begin{equation}
\bar{d} = \frac{1}{N}\sum_{i=1}^{N} d_i.
\label{eq:spacing}
\end{equation}

Both downsampled synthetic datasets were subsequently used independently to train our models. We applied a series of rotations to each plot to augment this data and increase its diversity. Specifically, we rotated each plot around the z-axis at six angles: 45, 90, 135, 180, 225, 270, and 315 degrees, resulting in additional plots. This process of systematic rotation allowed us to create an additional 8400  plots from the original 1200, thereby expanding the dataset to a total of 9600 plots. During the training process, the data was augmented dynamically at the start of each epoch using jittering techniques to introduce slight random variations, thereby enhancing the model's robustness to noise. This on-the-fly augmentation and rotation contributed to more effective model generalization and performance.

\subsection{Deep learning models for regression}

We developed four deep regression models based on commonly used point cloud encoder networks, namely, PointNet \citep{qi2017pointnet}, PointNet++ \citep{qi2017pointnet++}, Dynamic Graph Convolutional Neural Network (DGCNN) \citep{wang2019dynamic}, and PointConv \citep{wu2019pointconv}. These networks are well-suited for processing unstructured 3D point cloud data and have demonstrated effectiveness in capturing intricate spatial relationships within the data, thereby improving the accuracy of regression tasks. To adapt these models for regression, we modified the final classification layer by adjusting the output to predict continuous values, enabling accurate wood volume estimation.

\subsubsection{PointNet}

PointNet is a pioneering neural network architecture designed for processing 3D point clouds directly, bypassing the need for volumetric or mesh representations.  The primary innovation of PointNet lies in its ability to operate directly on point clouds, which are sets of points in a three-dimensional coordinate system, representing the external surfaces of objects. It tackles the challenges of unstructured point cloud data by using transformation networks (T-Nets) and a symmetric function, specifically max pooling, to maintain invariance to permutation. The T-Nets, including a spatial transformer network (STN), align the point cloud spatially and in feature space, enhancing the model's performance by learning optimal orientation for processing.  This architecture effectively aggregates and utilizes global features from the entire point cloud, ensuring robustness against disorder and transformational variances. PointNet provides substantial advancements in the field of computer vision and 3D object recognition.

\subsubsection{PointNet++}
PointNet++ is an advanced neural network architecture that extends the foundational concepts introduced by PointNet to address its limitations in capturing fine local structures within point clouds. It enhances the ability to process 3D point clouds by incorporating hierarchical neural networks that can capture detailed local features at multiple scales. This improvement is critical for tasks that require a nuanced understanding of complex geometric structures, such as detailed object segmentation and high-precision 3D recognition.

\subsubsection{DGCNN}
Dynamic Graph Convolutional Neural Network (DGCNN) further extends the capabilities of point cloud processing by dynamically updating the graph structure that represents the point cloud during training. It employs edge convolution operations to capture both local and global relationships between points, enabling the model to learn more intricate geometric features. The edge convolution layer functions by taking a tensor with dimensions N×F (where N is the number of points and F is the feature dimension of the input clouds) and applying it across various network layers. This layer dynamically constructs a graph for each point based on its K nearest neighbors and then uses this graph to compute new feature representations that capture both local structures and global context. The process involves creating K×N×M features (where M is the number of output classes or features), which are then aggregated using a pooling operation to produce an N×M dimensional output. This innovative method allows DGCNN to effectively generate new tensors that reflect the complex interrelationships within the data, significantly enhancing its capability for tasks such as classification and segmentation in 3D spaces.

\subsubsection{PointConv}

PointConv is a sophisticated convolutional operation designed for deep convolutional neural networks that process irregular and unordered point clouds. It uniquely handles the input coordinates of point clouds by utilizing a dynamic filter approach adapted for non-uniform sampling. In PointConv, the convolution weights are not static but are learned through a multi-layer perceptron (MLP). Additionally, it incorporates kernel density estimation to manage the density functions, ensuring the network can effectively adapt to non-uniform sampling conditions. This innovative approach allows PointConv to maintain scalability and robustness, offering translation invariance and permutation invariance in processing point clouds, making it highly effective for a range of applications in 3D data interpretation.

\subsection{Model training and  evaluation}


All four deep regression networks were implemented in Python using PyTorch \citep{paszke2019pytorch} and trained on NVIDIA A40 GPUs with 24\,GB of VRAM. Table~\ref{tab:hparams} summarizes the training hyperparameters used in all experiments. 

\begin{table}[t]
\centering
\caption{Training hyperparameters used for all deep regression models.}
\label{tab:hparams}
\renewcommand{\arraystretch}{1.05}
\setlength{\tabcolsep}{4pt}
\footnotesize
\begin{tabular}{p{0.62\columnwidth}p{0.30\columnwidth}}
\hline
\textbf{Hyperparameter} & \textbf{Value} \\
\hline
Initial learning rate & 0.001 \\
Batch size & 16 \\
Number of epochs & 200 \\
Optimizer & AdamW \\
Weight decay & $1\times10^{-4}$ \\
Learning rate scheduler & Cosine annealing warm restarts \citep{loshchilov2016sgdr} \\
$T_{0}$ (first restart) & 200 steps \\
$T_{\text{mult}}$ & 1 \\
$\eta_{\min}$ & $1\times10^{-6}$ \\
\hline
\end{tabular}
\end{table}

Our training dataset included synthetic point clouds of 9600 plots. Using 5-fold cross-validation, the dataset was divided into 5 equal parts. In each iteration, 7680 plots (4 folds) were used for training, while the remaining 1920 (1 fold) were used for validation. This process was repeated 5 times, ensuring that each fold was used once for validation, providing a robust evaluation of the model's performance. The training process minimizes the prediction error defined by the loss function. We used the Mean Squared Error (MSE) as the loss function during training:

\begin{equation}
\mathcal{L} =\frac{1}{n} \sum_{i=1}^{n} (y_i - \hat{y}_i)^2
\label{eq:Loss}
\end{equation}

where $n$ is the total number of samples, $y_i$ is the ground truth volume for the $i$th sample, and $\hat{y}_i$ is the predicted volume for the $i$th sample.

For the evaluation of the regression results on validation and test samples we used the Mean Absolute Percentage Error (MAPE) as the evaluation metric:

\begin{equation}
\text{MAPE} = \frac{1}{n} \sum_{i=1}^{n} \left| \frac{y_i - \hat{y}_i}{y_i} \right| \times 100
\label{eq:MAPE}
\end{equation}

The MAPE measures the prediction error relative to the ground truth volume, which makes it easier to interpret the performance of the models across different scales. The use of the 5-fold cross validation enables us to compute both the mean and standard deviation (STD) of predictions across 5 folds. The mean value, calculated over the five folds, provides an average performance metric for each model. The standard deviation reflects the variability in performance across the folds, with lower STD values indicating greater consistency and stability across the different training/validation splits. 

After each model was trained on synthetic data, we used it to predict the wood volume, AGB, and carbon of real point cloud data. To estimate wood volume from the real point cloud data, the large plot was subdivided into smaller, discrete tiles, similar to the synthetic plots, which served as inputs for the trained models. Prior to model inference, the point cloud data for each tile was downsampled to 2048 points using the same methods, random sampling and farthest point sampling, which were applied to the training data. We converted the predicted wood volume for each tile to AGB by applying species-specific wood density values, and then further converted it to carbon using a factor of 0.5, facilitating a robust estimation of carbon from volume data.

\subsection{Allometric AGB estimation}

To compare our direct AGB estimation results against indirect (allometry-based) baselines, we implemented three state-of-the-art deep-learning individual tree segmentation baselines: SegmentAnyTree, TreeLearn, and ForestFormer3D.
We converted the resulting tree instances to AGB using a general Eucalyptus allometric model for estimating AGB from tree height. After segmenting the point cloud into individual trees, we computed a per-tree height metric (H) for every detected instance and used this as the predictor for AGB estimation. Tree height was derived from the ground-normalized point cloud as the vertical extent of each segmented instance (i.e., canopy top height relative to the local ground surface). Using the estimated height, AGB for each individual tree was calculated with the allometric equation \citep{williams2005allometry} given below:

\begin{equation}
\label{generalalmodel}
\mathrm{AGB} = \exp(-3.5413 + 3.5337 \ln(H)) 
\end{equation}

where H is in meters and AGB is predicted per tree (kg). Plot-level AGB was then obtained by summing individual-tree AGB across all detected trees within each plot boundary for each segmentation method. Additionally, we compared the performance of deep regression networks with an indirect CHM-segmentation approach for individual-tree delineation \citep{khoshelham2023tree}, as well as the FullCAM. In the CHM-segmentation method, individual trees were delineated using marker-controlled watershed segmentation \citep{chen2006isolating} and tree heights were extracted and converted to AGB using a above allometric model in Equation \ref{generalalmodel}. FullCAM, developed by the Australian Government, is an empirical model designed to estimate changes in carbon pools and greenhouse gas emissions across various large-scale vegetation types, including forests, agricultural areas, and other vegetated regions. The model predicts the accumulation of AGB in woody vegetation, which is then used to calculate carbon sequestration \citep{roxburgh2019revised}. The results are also validated against field measurements to determine which approach provides estimates that are most consistent with field data.

\section{Results}\label{sec4}

In this section, we compare the characteristics of the synthetic and real point cloud datasets. We then analyze the impact of downsampling on model performance and report results on synthetic data as well as on real point clouds. Furthermore, we benchmark the proposed direct approach against indirect AGB estimation baselines based on individual tree segmentation followed by allometric conversion, and against FullCAM, using field measurements as reference.

\subsection{Point clouds characteristic}
The characteristics of the downsampled synthetic and real point cloud datasets are summarized in Table \ref{tab:downsampling_comparison}. The synthetic dataset contains 9600 plots (2048 points per plot) and is used for training/validation via 5-fold cross-validation (\(K=5\)). The real dataset consists of five sites totaling 293 plots (2048 points per plot), held out for testing. Key metrics, such as plot area and point density, are detailed in the table, providing insights into the spatial structure and distribution of the datasets.
Across all six datasets, Farthest Point Sampling produces larger mean plot areas than Random Sampling, indicating broader spatial coverage: for the synthetic set, area means are \(300.32 \pm 50.38\ \text{m}^2\) (Random Sampling) versus \(315.15 \pm 52.51\ \text{m}^2\) (Farthest Point Sampling). For the five real sites, area means under Random Sampling range from \(277.85\)–\(339.92\ \text{m}^2\) and under Farthest Point Sampling from \(284.60\)–\(353.64\ \text{m}^2\); averaged across sites, the mean area under \emph{Farthest Point Sampling} is larger than under \emph{Random Sampling}.
Also, Farthest Point Sampling produces slightly lower point densities than Random Sampling. For the synthetic set, point density per \(\text{m}^2\) is \(7.03 \pm 1.28\) (Random Sampling) versus \(6.69 \pm 1.21\) (Farthest Point Sampling), and per \(\text{m}^3\) is \(0.59 \pm 0.14\) (Random Sampling) versus \(0.44 \pm 0.08\) (Farthest Point Sampling). Across the five real sites, mean point densities per \(\text{m}^2\) range from \(6.33\)–\(7.53\) under Random Sampling and \(6.12\)–\(7.32\) under Farthest Point Sampling. Similarly, mean densities per \(\text{m}^3\) range from \(0.41\)–\(0.50\) (Random Sampling) and \(0.38\)–\(0.48\) (Farthest Point Sampling). In addition, average point spacing is consistently larger under farthest point sampling than under random sampling across both synthetic and real datasets. Overall, Farthest Point Sampling yields lower point density than Random Sampling for both \(\text{m}^2\) and \(\text{m}^3\). Together, larger mean plot areas and larger point spacing in both synthetic and real datasets sampled using farthest point sampling indicate broader spatial coverage with less local clustering than random sampling. By prioritizing points that are farthest apart, farthest point sampling ensures broader spatial coverage, potentially capturing more diverse sections of the plot. In contrast, random sampling tends to preserve denser clusters, possibly leading to a smaller represented area and higher local densities. These patterns highlight the trade-offs between the two methods, with farthest point sampling promoting better generalization by emphasizing structural outlines and reducing clustering biases.

\newcommand{\vpm}[2]{\begin{tabular}{@{}c@{}}#1 \\ $\pm$ \\ #2\end{tabular}}

\begin{table*}
\caption{Comparison of downsampled dataset characteristics using random sampling (RS) and farthest point sampling (FPS) for synthetic and real point cloud datasets. Dataset ID: MF = Melville Forest, HP = Hensley Park, 1995.1 = area no. 1 in the 1995 planatation year}
\label{tab:downsampling_comparison}
\centering
\renewcommand{\arraystretch}{1.15}
\scriptsize 
\setlength\tabcolsep{3pt}
\begin{tabular}{lcccccccccccc}
\hline
\textbf{Metric} & \multicolumn{2}{c}{\textbf{Synthetic}} & \multicolumn{2}{c}{\textbf{MF1995.1}} & \multicolumn{2}{c}{\textbf{MF1999.2}} & \multicolumn{2}{c}{\textbf{MF2007.2}} & \multicolumn{2}{c}{\textbf{HP2010.2}} & \multicolumn{2}{c}{\textbf{Knewleaves20}} \\
\cline{2-3} \cline{4-5} \cline{6-7} \cline{8-9} \cline{10-11} \cline{12-13}
& \textbf{RS} & \textbf{FPS} & \textbf{RS} & \textbf{FPS} & \textbf{RS} & \textbf{FPS} & \textbf{RS} & \textbf{FPS} & \textbf{RS} & \textbf{FPS} & \textbf{RS} & \textbf{FPS} \\
\hline
Plot area (m$^{2}$) 
& \vpm{300.32}{50.38} & \vpm{315.15}{52.51} 
& \vpm{301.74}{53.89} & \vpm{313.33}{55.40} 
& \vpm{279.73}{44.02} & \vpm{288.20}{44.99} 
& \vpm{339.92}{69.24} & \vpm{353.64}{75.94} 
& \vpm{277.85}{38.52} & \vpm{284.60}{39.26} 
& \vpm{335.20}{49.72} & \vpm{341.85}{42.10} \\
\addlinespace[3pt]
Point density (m$^{-2}$) 
& \vpm{7.03}{1.28} & \vpm{6.69}{1.21} 
& \vpm{7.05}{1.53} & \vpm{6.79}{1.48} 
& \vpm{7.53}{1.37} & \vpm{7.30}{1.32} 
& \vpm{6.45}{2.35} & \vpm{6.23}{2.30} 
& \vpm{7.50}{0.97} & \vpm{7.32}{0.94} 
& \vpm{6.33}{1.58} & \vpm{6.12}{1.07} \\
\addlinespace[3pt]
Point density (m$^{-3}$) 
& \vpm{0.59}{0.14} & \vpm{0.45}{0.08} 
& \vpm{0.45}{0.12} & \vpm{0.42}{0.12} 
& \vpm{0.50}{0.15} & \vpm{0.47}{0.14} 
& \vpm{0.46}{0.30} & \vpm{0.43}{0.28} 
& \vpm{0.50}{0.13} & \vpm{0.48}{0.12} 
& \vpm{0.41}{0.17} & \vpm{0.38}{0.13} \\

Average point spacing (cm) 
& \vpm{0.33}{0.03} & \vpm{0.71}{0.06}
& \vpm{0.26}{0.04} & \vpm{0.59}{0.10} 
& \vpm{0.28}{0.04} & \vpm{0.60}{0.10} 
& \vpm{0.32}{0.05} & \vpm{0.72}{0.12} 
& \vpm{0.32}{0.04} & \vpm{0.71}{0.10} 
& \vpm{0.41}{0.08} & \vpm{1}{0.19} \\

\hline
\end{tabular}
\end{table*}

\subsection{Model performance  on synthetic data}

The training and validation performance of the four deep learning models was evaluated using 5-fold cross-validation on synthetic datasets downsampled with both the farthest point sampling and random sampling methods. 

\begin{figure*}[!t]
  \centering
  \begin{subfigure}[b]{0.45\linewidth}
    \centering
    \includegraphics[width=\textwidth]{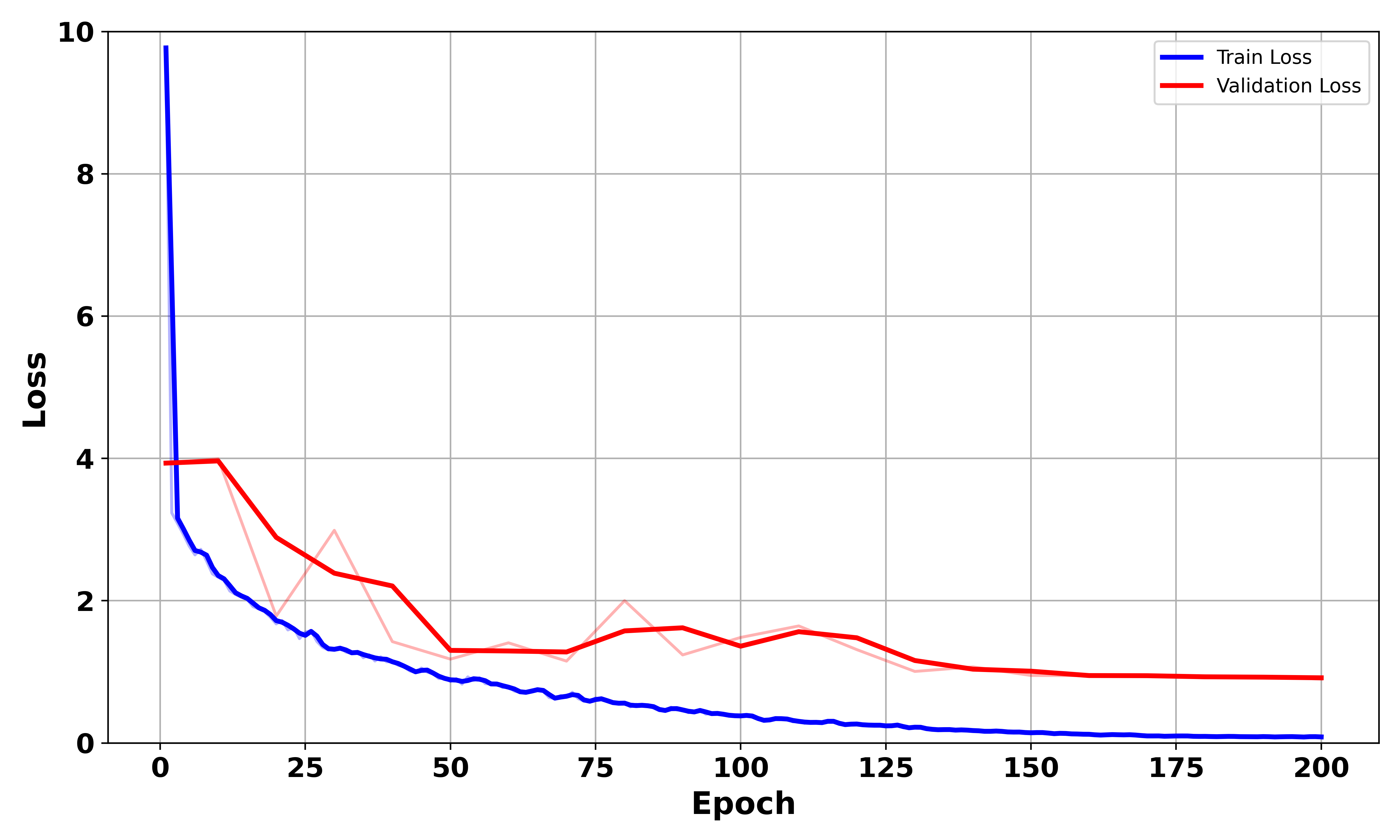}
    \caption{}
    \label{fig:pnet}
  \end{subfigure}
  \hfill 
  \begin{subfigure}[b]{0.45\linewidth}
    \centering
    \includegraphics[width=\textwidth]{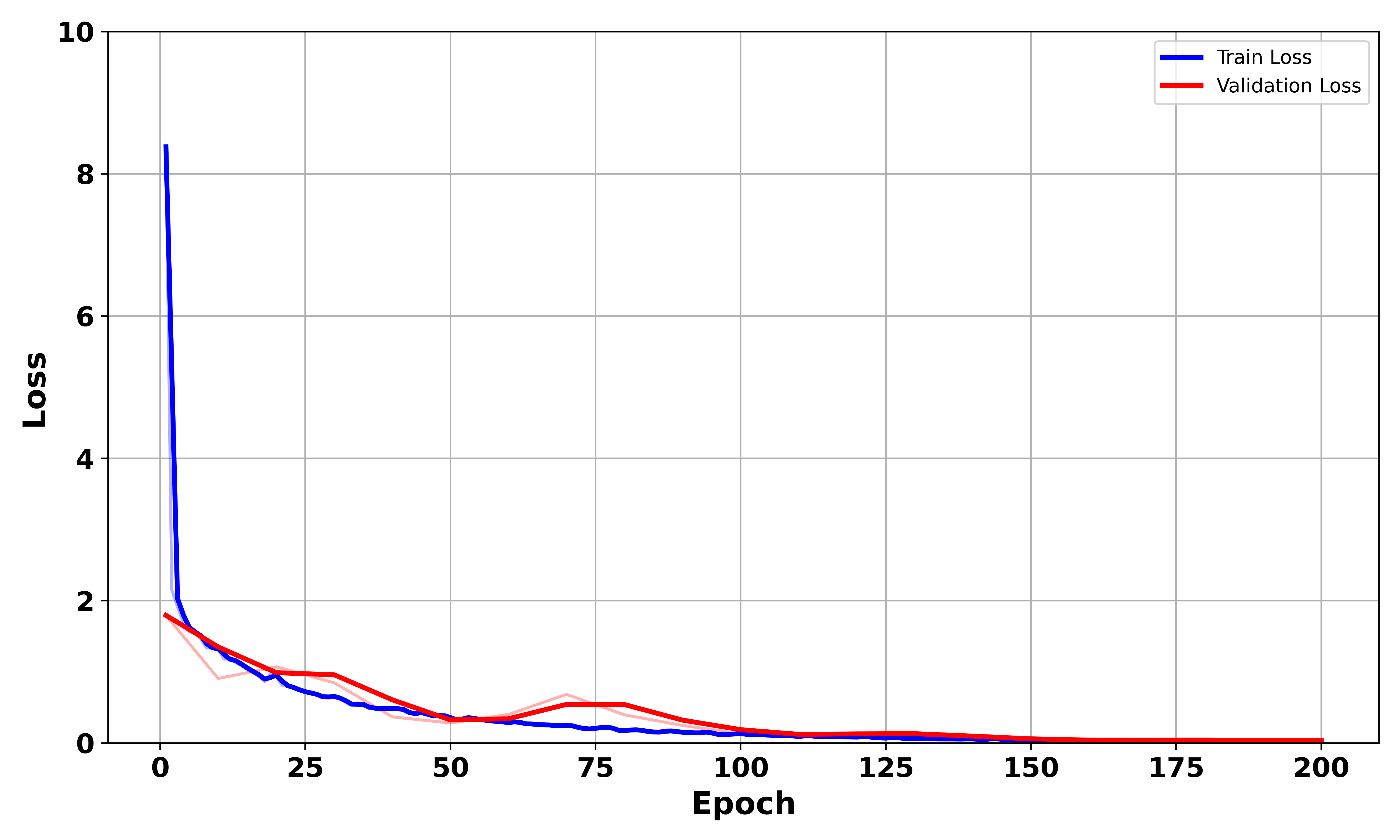}
    \caption{}
    \label{fig:pnet2}
  \end{subfigure}
    
  \vspace{1.5em} 

  \begin{subfigure}[b]{0.45\linewidth}
    \centering
    \includegraphics[width=\textwidth]{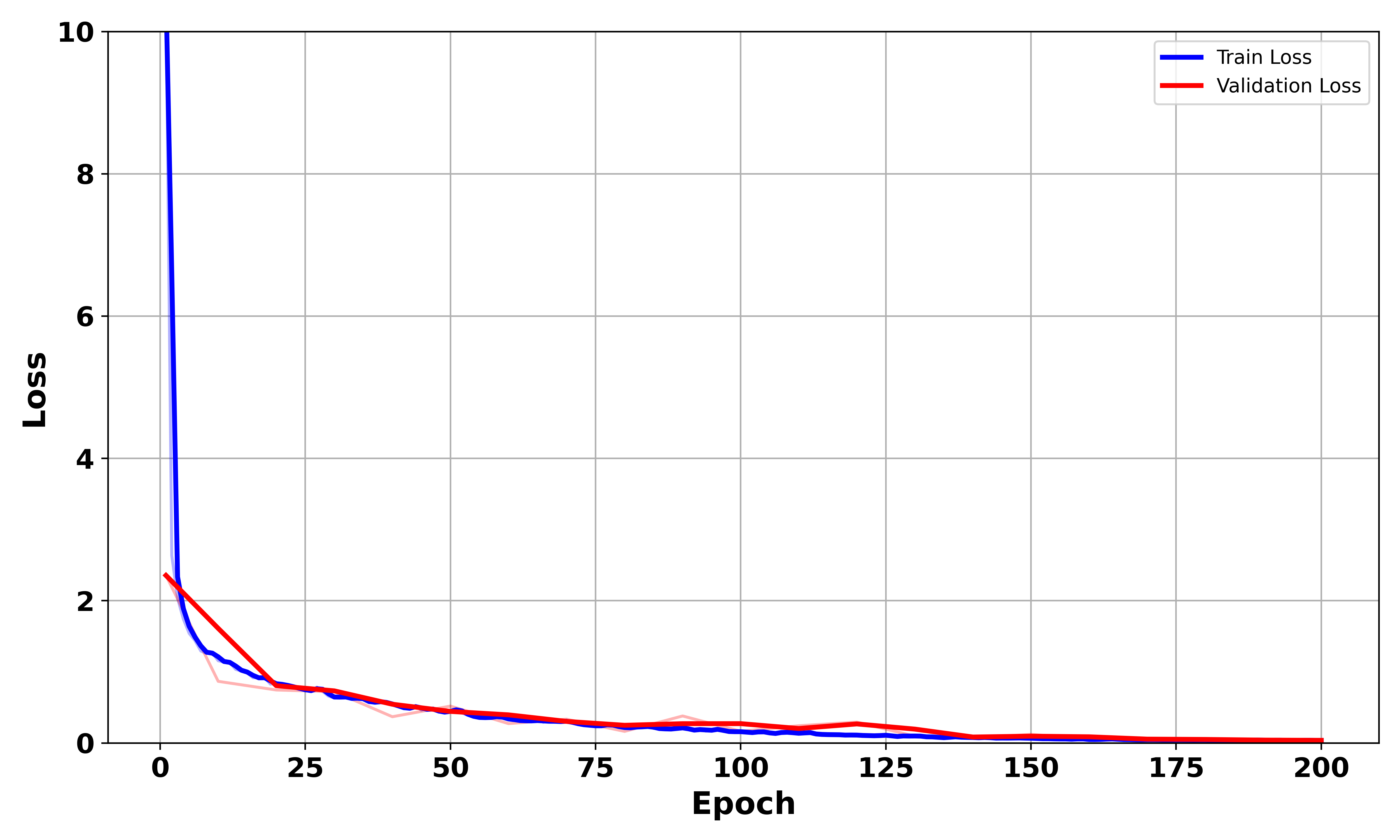}
    \caption{}
    \label{fig:dgcnn}
  \end{subfigure}
  \hfill
  \begin{subfigure}[b]{0.45\linewidth}
    \centering
    \includegraphics[width=\textwidth]{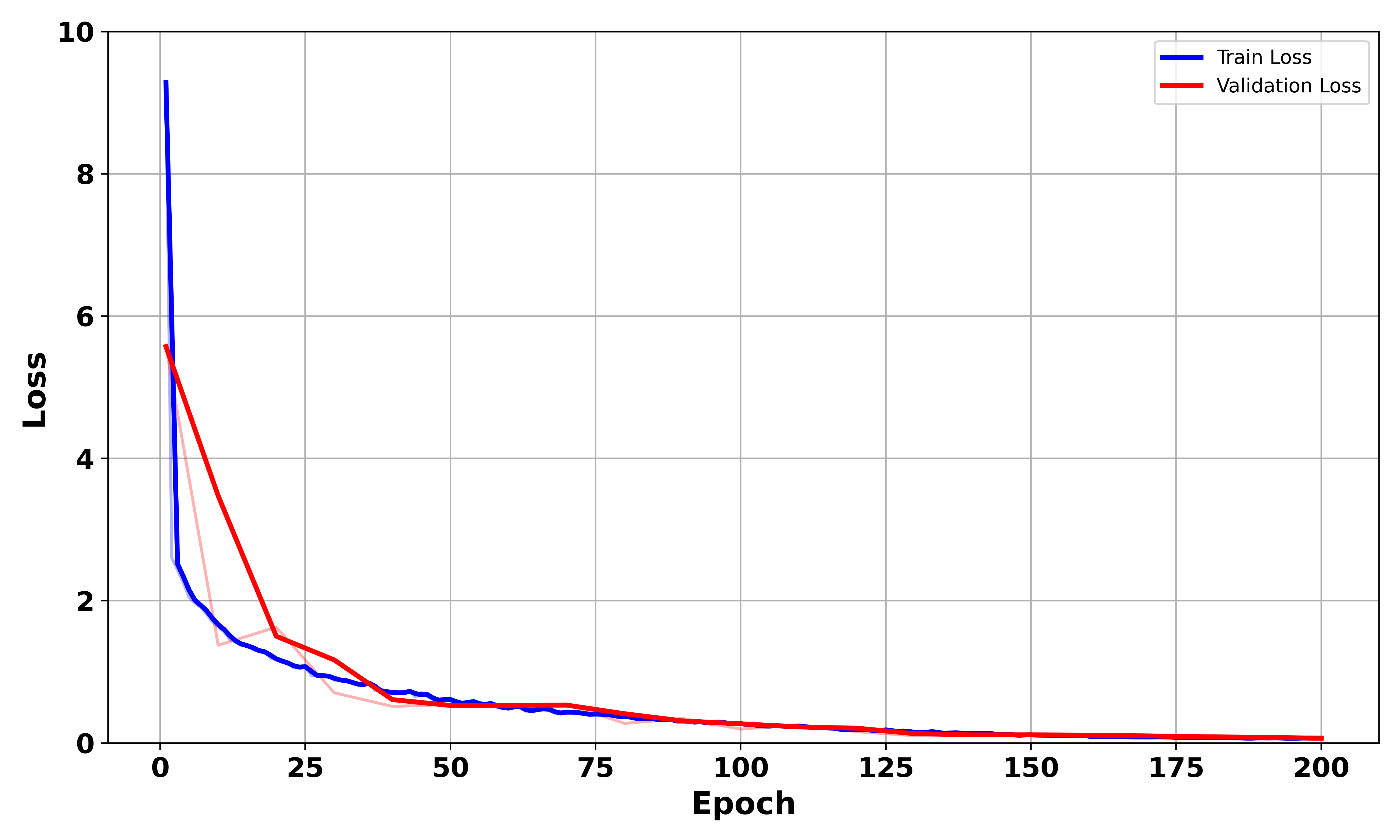}
    \caption{}
    \label{fig:pconv}
  \end{subfigure}

  \caption{Learning curves of the deep learning models trained on synthetic point cloud data downsampled using the farthest point sampling method: (a) PointNet, (b) PointNet++, (c) DGCNN, and (d) PointConv.}
  \label{fig:learnningcurve}
\end{figure*}

Figure \ref{fig:learnningcurve} shows the loss curves for each model over 200 epochs, highlighting their learning and generalization capabilities. PointNet exhibited a steady decline in training loss; however, its validation loss fluctuated significantly, especially during the initial epochs. This indicates that while PointNet effectively learns the training data, it struggles with generalization, making it less reliable for unseen data. In contrast, the other models  displayed more  consistent reductions in both training and validation losses, demonstrating strong generalization ability and minimal overfitting.
Overall, PointNet++ and PointConv demonstrated the most robust and stable learning behavior, with DGCNN performing nearly as well despite minor early-stage variability. PointNet, however, faced challenges in achieving generalization stability, limiting its effectiveness.

Table \ref{tab:trainresults} presents the validation error and MAPE for all models based on 5-fold cross-validation. These results were obtained by training the models on synthetic data, which was downsampled using two different methods: random sampling and farthest point sampling. The values in Table \ref{tab:trainresults} were calculated by averaging the results across all 5 folds, ensuring robust generalization across the dataset. PointNet exhibited the highest error rates for random sampling, with a validation error of 1.034 \(\pm 0.019\) and a MAPE of 8.999 \(\pm 0.189\%\), while farthest point sampling slightly improved these metrics, achieving a validation error of 0.959 \(\pm 0.029\) and a MAPE of 8.111 \(\pm 0.204\%\). These results highlight the limited ability of PointNet to generalize effectively on synthetic data, likely due to its lack of neighborhood-awareness in processing point cloud data.
PointNet++ emerged as the best-performing model, achieving the lowest validation error across both downsampling methods 
and the lowest MAPE with random sampling. 
DGCNN and 
PointConv also showed strong performance, with validation error 
and MAPE values very close to PointNet++. 
In summary, both random sampling and farthest point sampling methods yielded comparable performances across all models, with PointNet++ and DGCNN demonstrating the most reliable results for wood volume estimation on synthetic data. PointConv also performed well, while PointNet's higher error rates and variability underscore the importance of incorporating neighborhood information for improved generalization in point cloud processing tasks. 

\begin{table*}[!t]
\caption{Mean \(\pm\) STD of validation error and MAPE for PointNet, PointNet++, DGCNN, and PointConv trained on synthetic datasets, based on 5-fold cross-validation}
\label{tab:trainresults}
\centering
\renewcommand{\arraystretch}{1.2}
\footnotesize
\begin{tabular}{lcccc}
\hline
\textbf{Model} & \multicolumn{2}{c}{\textbf{Random Sampling}} & \multicolumn{2}{c}{\textbf{Farthest Point Sampling}} \\
\cline{2-3} \cline{4-5}
 & \textbf{Validation Error (m$^3$)} & \textbf{MAPE (\%)} & \textbf{Validation Error (m$^3$)} & \textbf{MAPE (\%)} \\
\hline
PointNet   & 1.034 $\pm$ 0.019 & 8.999 $\pm$ 0.189 & 0.959 $\pm$ 0.029 & 8.111 $\pm$ 0.204 \\
PointNet++ & 0.151 $\pm$ 0.016 & 1.438 $\pm$ 0.178 & 0.190 $\pm$ 0.010 & 1.797 $\pm$ 0.149 \\
DGCNN      & 0.195 $\pm$ 0.010 & 1.627 $\pm$ 0.060 & 0.202 $\pm$ 0.015 & 1.691 $\pm$ 0.146 \\
PointConv  & 0.202 $\pm$ 0.019 & 1.705 $\pm$ 0.138 & 0.286 $\pm$ 0.040 & 2.307 $\pm$ 0.254 \\
\hline
\end{tabular}
\end{table*}

\subsection{Model performance on real data}
After training the deep learning models on synthetic data, we evaluated their performance on real point cloud datasets to estimate wood volume. The estimated wood volumes were converted to AGB using species-specific wood density values derived from the study by Young \cite{young2000variation}, which focused on eucalyptus trees. Furthermore, a conversion factor of 0.5 was applied to the AGB estimates to calculate the corresponding carbon stock values. Tables \ref{tab:fps_agbc} and \ref{tab:rs_agbc} presents the AGB (t/ha) and
carbon stock (t/ha) estimated by the deep regression networks (PointNet, PointNet++, DGCNN, and PointConv) applied to real point cloud datasets. 

\begin{table*}[!t]
\caption{AGB and carbon (t/ha) from real point clouds downsampled with fathest point sampling for Jigsaw Farm (MF1995.1, MF1999.2, MF2007.2, HP2010.2) and Knewleaves Farm (Knewleave20)}
\label{tab:fps_agbc}
\centering
\renewcommand{\arraystretch}{1.2}
\footnotesize
\setlength\tabcolsep{4pt}
\begin{tabular}{lcccccccccc}
\hline
\textbf{Model} & \multicolumn{2}{c}{\textbf{MF1995.1}} & \multicolumn{2}{c}{\textbf{MF1999.2}} & \multicolumn{2}{c}{\textbf{MF2007.2}} & \multicolumn{2}{c}{\textbf{HP2010.2}} & \multicolumn{2}{c}{\textbf{Knewleave20}} \\
\cline{2-3} \cline{4-5} \cline{6-7} \cline{8-9} \cline{10-11}
& \textbf{AGB (t/ha)} & \textbf{C (t/ha)} 
& \textbf{AGB (t/ha)} & \textbf{C (t/ha)} 
& \textbf{AGB (t/ha)} & \textbf{C (t/ha)} 
& \textbf{AGB (t/ha)} & \textbf{C (t/ha)} 
& \textbf{AGB (t/ha)} & \textbf{C (t/ha)} \\
\hline
PointNet   & 215 $\pm$ 20 & 107 $\pm$ 10 & 279 $\pm$ 30 & 139 $\pm$ 15 & 195 $\pm$ 13 & 97 $\pm$ 7 & 277 $\pm$ 34 & 139 $\pm$ 17 & 206 $\pm$ 27 & 103 $\pm$ 14 \\
PointNet++ & 200 $\pm$ 2  & 100 $\pm$ 1  & 237 $\pm$ 2  & 119 $\pm$ 1  & 181 $\pm$ 1  & 91 $\pm$ 1 & 225 $\pm$ 4  & 113 $\pm$ 2  & 205 $\pm$ 7  & 103 $\pm$ 3 \\
DGCNN      & 154 $\pm$ 12 & 77 $\pm$ 6   & 200 $\pm$ 12 & 100 $\pm$ 6  & 149 $\pm$ 8  & 75 $\pm$ 4 & 205 $\pm$ 15 & 103 $\pm$ 7  & 170 $\pm$ 11 & 85 $\pm$ 5 \\
PointConv  & 186 $\pm$ 26 & 93 $\pm$ 13  & 201 $\pm$ 17 & 100 $\pm$ 9  & 150 $\pm$ 18 & 75 $\pm$ 9 & 188 $\pm$ 28 & 94 $\pm$ 14  & 222 $\pm$ 89 & 111 $\pm$ 44 \\
\hline
\end{tabular}
\end{table*}

\begin{table*}[!t]
\caption{AGB and carbon (t/ha) from real point clouds downsampled with random sampling for Jigsaw Farm (MF1995.1, MF1999.2, MF2007.2, HP2010.2) and Knewleaves Farm (Knewleave20)}
\label{tab:rs_agbc}
\centering
\renewcommand{\arraystretch}{1.2}
\footnotesize
\setlength\tabcolsep{4pt}
\begin{tabular}{lcccccccccc}
\hline
\textbf{Model} & \multicolumn{2}{c}{\textbf{MF1995.1}} & \multicolumn{2}{c}{\textbf{MF1999.2}} & \multicolumn{2}{c}{\textbf{MF2007.2}} & \multicolumn{2}{c}{\textbf{HP2010.2}} & \multicolumn{2}{c}{\textbf{Knewleave20}} \\
\cline{2-3} \cline{4-5} \cline{6-7} \cline{8-9} \cline{10-11}
& \textbf{AGB (t/ha)} & \textbf{C (t/ha)}
& \textbf{AGB (t/ha)} & \textbf{C (t/ha)}
& \textbf{AGB (t/ha)} & \textbf{C (t/ha)}
& \textbf{AGB (t/ha)} & \textbf{C (t/ha)}
& \textbf{AGB (t/ha)} & \textbf{C (t/ha)} \\
\hline
PointNet   & 130 $\pm$ 15 & 65 $\pm$ 7  & 193 $\pm$ 33 & 97 $\pm$ 16  & 125 $\pm$ 19 & 63 $\pm$ 10  & 179 $\pm$ 37 & 89 $\pm$ 19  & 114 $\pm$ 5  & 57 $\pm$ 3 \\
PointNet++ & 135 $\pm$ 2  & 67 $\pm$ 1  & 188 $\pm$ 3  & 94 $\pm$ 2   & 119 $\pm$ 1  & 60 $\pm$ 1   & 170 $\pm$ 4  & 85 $\pm$ 2   & 113 $\pm$ 2  & 57 $\pm$ 1 \\
DGCNN      & 125 $\pm$ 5  & 63 $\pm$ 3  & 170 $\pm$ 9  & 85 $\pm$ 4   & 112 $\pm$ 5  & 56 $\pm$ 3   & 158 $\pm$ 8  & 79 $\pm$ 4   & 104 $\pm$ 4  & 52 $\pm$ 2 \\
PointConv  & 130 $\pm$ 7  & 65 $\pm$ 3  & 181 $\pm$ 8  & 91 $\pm$ 4   & 117 $\pm$ 5  & 58 $\pm$ 2   & 164 $\pm$ 11 & 82 $\pm$ 6   & 115 $\pm$ 4  & 58 $\pm$ 2 \\
\hline
\end{tabular}
\end{table*}

As shown in tables \ref{tab:fps_agbc} and \ref{tab:rs_agbc}, among the deep regression models, PointNet++ demonstrated the most consistent performance as indicated by smaller standard deviation across both downsampling methods. For example, in the Melville Forest 2007.2, it yielded an AGB estimate of 181 ± 1 t/ha using farthest point sampling and 119 ± 1 t/ha with random sampling.
The low standard deviations highlight the ability of PointNet++ to provide consistent predictions with minimal variability.
DGCNN produced the lowest mean wood volume estimates among all models across both downsampling methods. 
Across downsampling strategies, farthest point sampling consistently produced higher AGB estimates than random sampling. These estimates were generally closer to the field measurements, suggesting that farthest point sampling can reduce the underestimation bias observed with random sampling when transferring from synthetic training data to real point clouds. However, the magnitude of these differences in some plots indicates that the predictions are sensitive to the downsampling/preprocessing choice, and this sensitivity should be considered when interpreting AGB and wood volume estimates.

\subsection{Effect of downsampling strategies on model performance}

\begin{figure}
\centering
\includegraphics[width=3.3in]{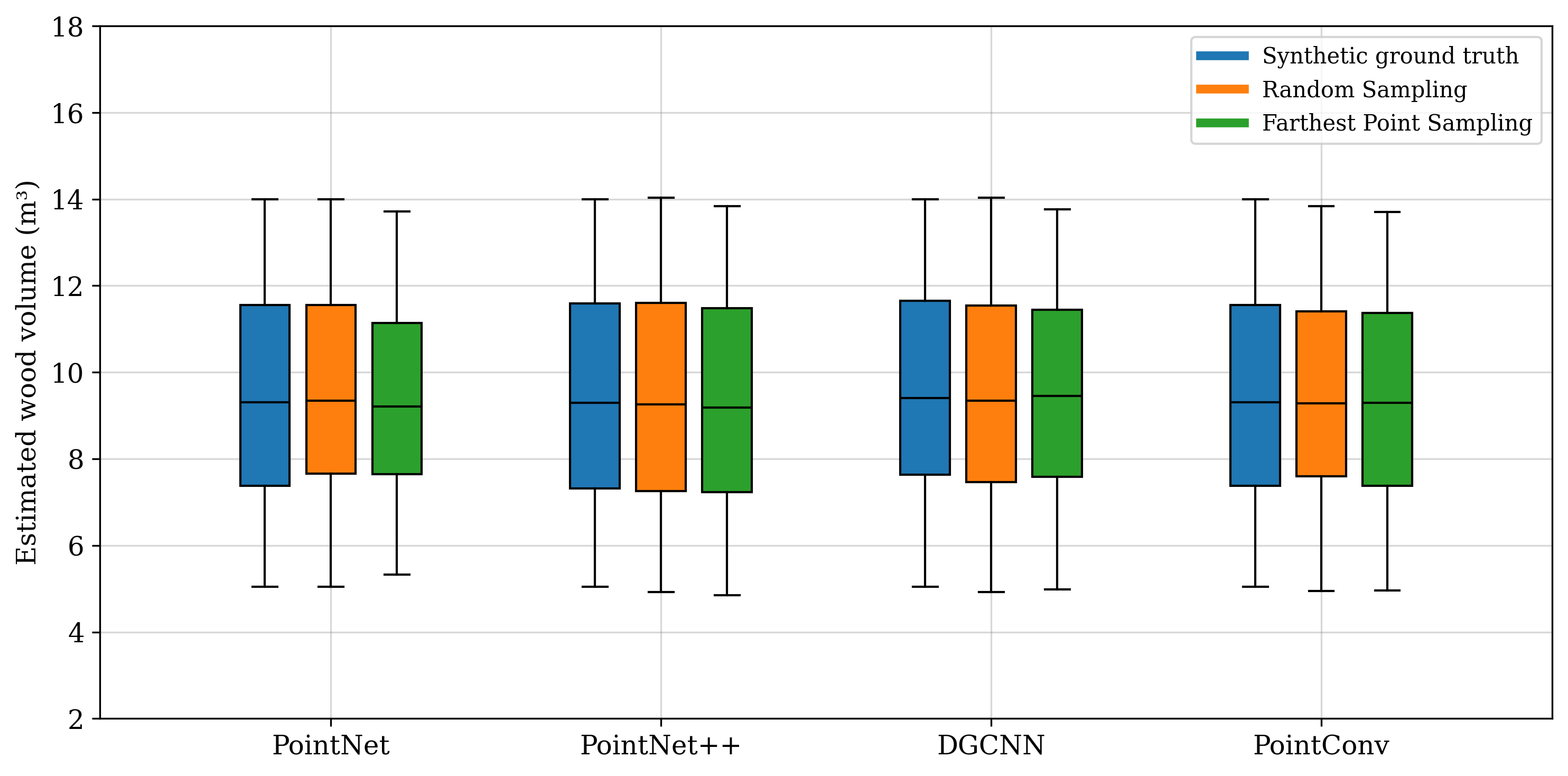}
\caption{Distribution of wood volume across synthetic plots estimated by the four models trained on synthetic data  downsampled using random sampling and farthest point sampling methods.}
\label{fig:fpsvsfxsynth}
\end{figure}

In this section, we investigate how different downsampling strategies, random sampling and farthest point sampling, affect model performance when trained on synthetic datasets and applied to real point cloud data. Figure \ref{fig:fpsvsfxsynth} illustrates the distribution of wood volume across the synthetic plots based on the validation results for  the four deep learning models such as PointNet, PointNet++, DGCNN, and PointConv. 
The results shown correspond to the best run out of the 5-fold cross-validation, selected based on the lowest validation loss. Across all models, the distributions of predicted wood volumes align closely with the synthetic ground truth, demonstrating the effectiveness of the training process. Both random sampling and farthest point sampling methods yield comparable validation results, with overlapping interquartile ranges and medians across all models. This consistency suggests that both downsampling techniques preserved the essential spatial characteristics required for model generalization within the synthetic dataset. Notably, all models show minimal variability, indicating stability in predictions across validation plots, with no significant discrepancies in performance among the architectures when applied to synthetic data. These findings highlight the reliability of the models and the ability of both downsampling techniques to maintain prediction accuracy during validation on synthetic datasets.

\begin{figure}
\centering
\includegraphics[width=3.3in]{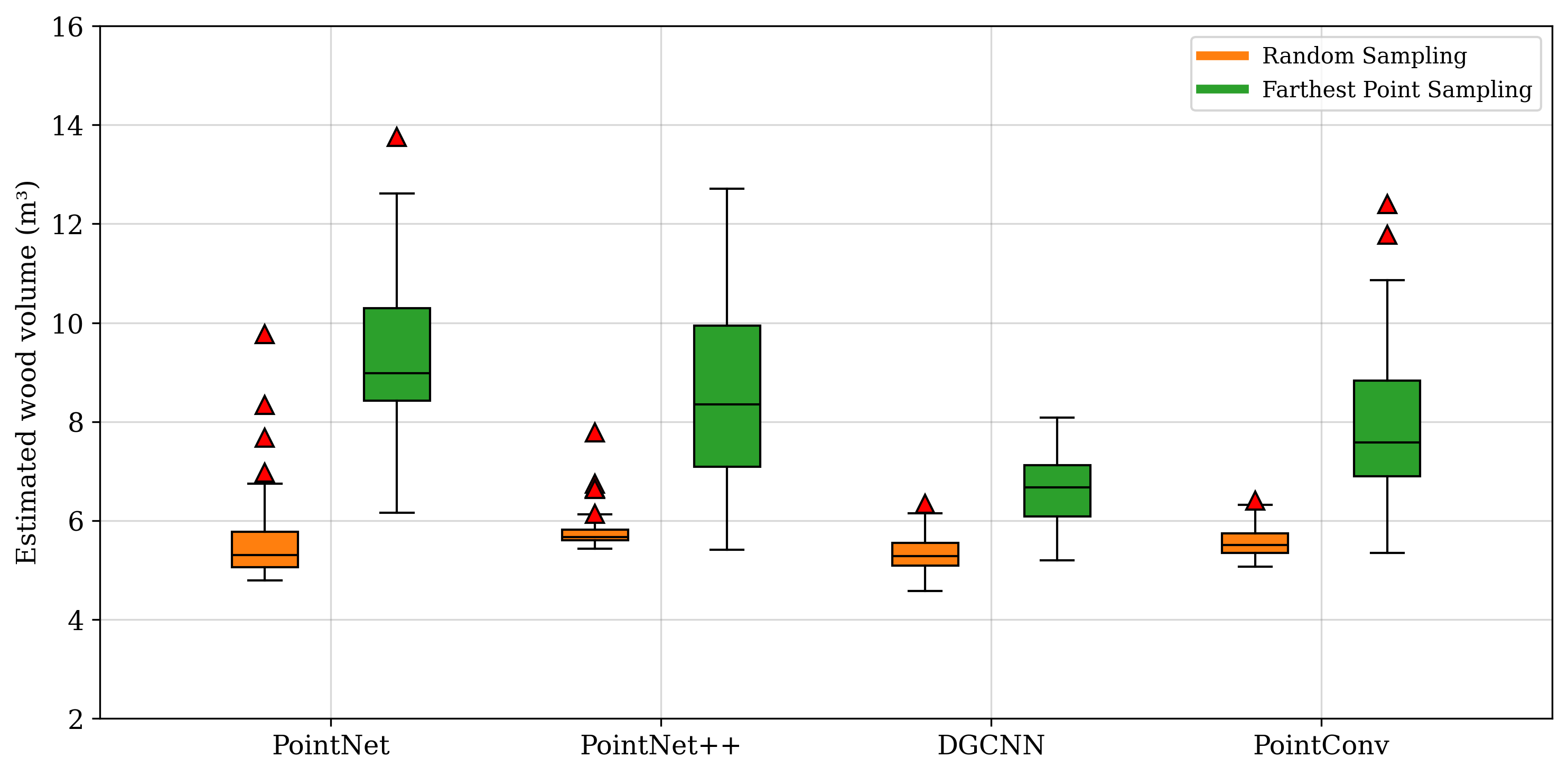}
\caption{The distribution of wood volume across real point cloud dataset in the MF1995.1, as estimated by the four models trained on synthetic data downsampled using random sampling and farthest point sampling methods.}
\label{fig:fpsvsfxreal}
\end{figure}

Figure \ref{fig:fpsvsfxreal} illustrates the distribution of wood volume estimated at the  Melville Forest 1995.1 by the four deep learning models trained on the synthetic datasets. The results demonstrate notable differences in the performance of the models when applied to real data. Across all models, farthest point sampling consistently yielded higher wood volume estimates with larger interquartile ranges than random sampling, indicating that farthest point sampling better preserved the spatial representation of the data and enabled improved generalization to real point cloud datasets. In contrast, random sampling resulted in significantly underestimated wood volumes across all models, with narrower interquartile ranges. 

\begin{figure*}[!t]
\centering
  \begin{subfigure}[b]{0.22\linewidth}
    \centering
    \includegraphics[width=\textwidth]{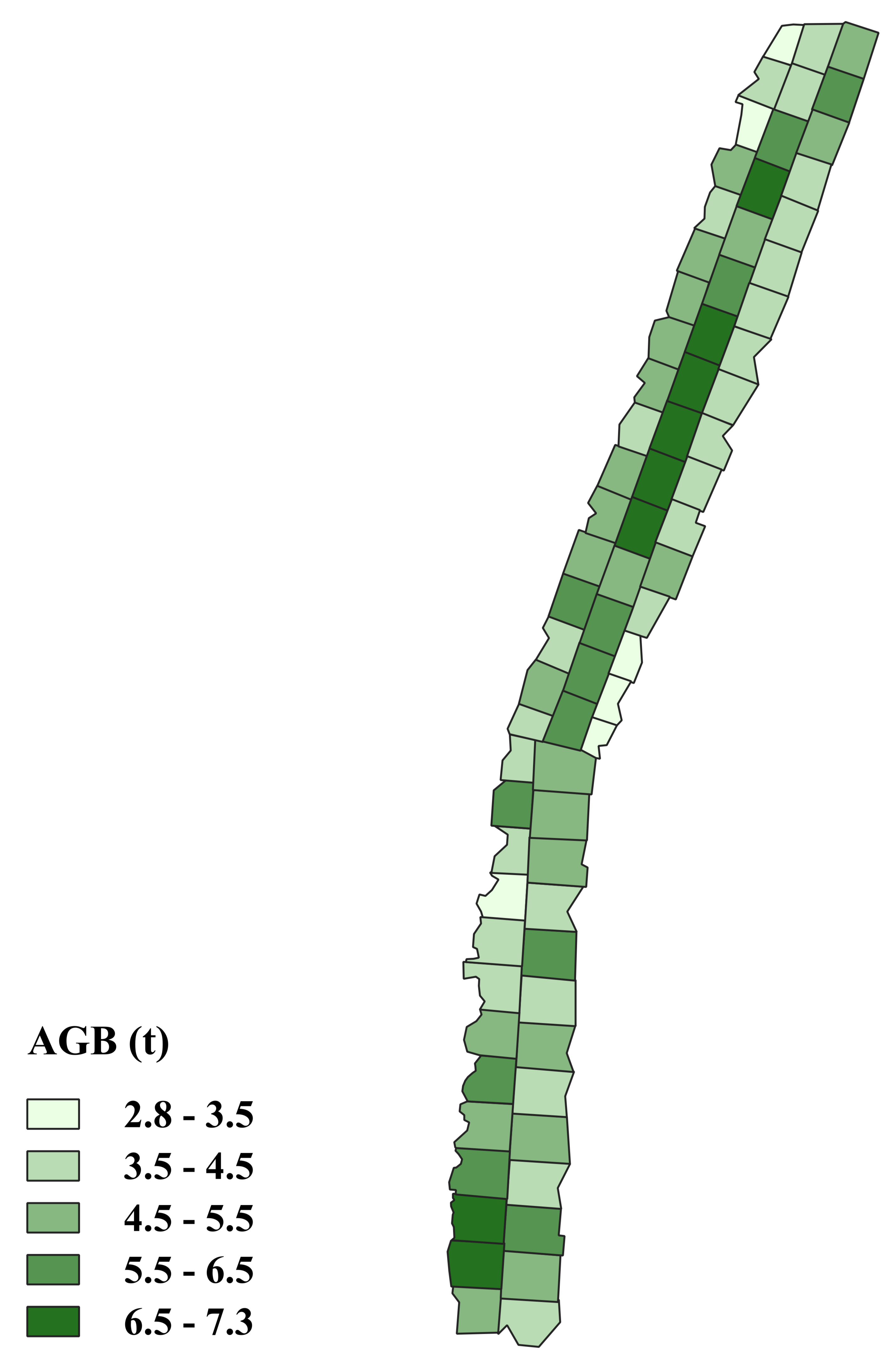}
    \caption{}
    \label{fig:19951}
  \end{subfigure}
  \hfill 
  \begin{subfigure}[b]{0.10\linewidth}
    \centering
    \includegraphics[width=\textwidth]{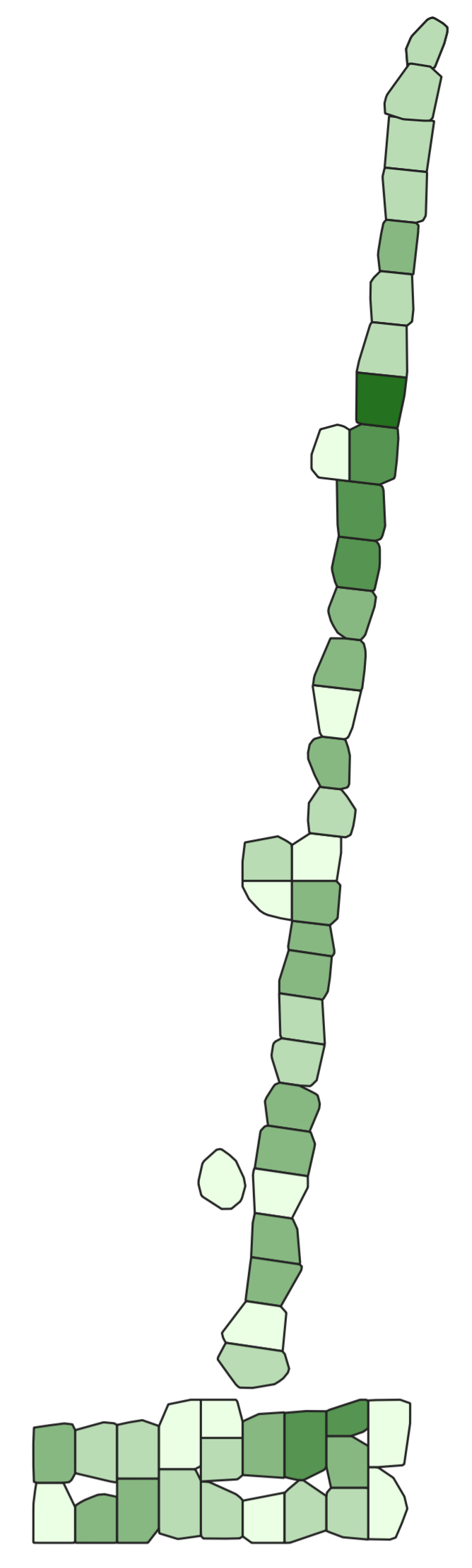}
    \caption{}
    \label{fig:19922}
  \end{subfigure}
  \hfill
  \begin{subfigure}[b]{0.25\linewidth}
    \centering
    \includegraphics[width=\textwidth]{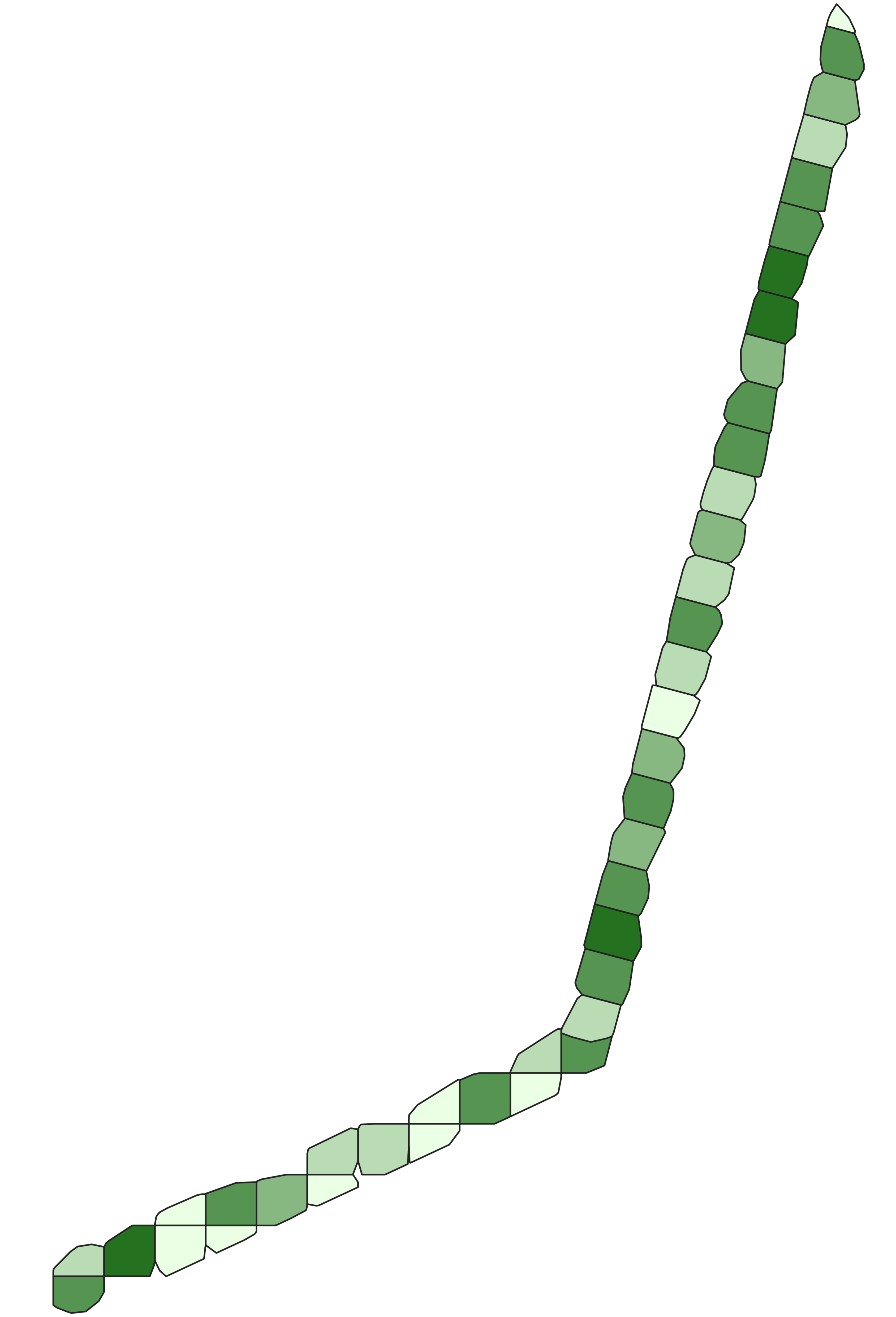}
    \caption{}
    \label{fig:20072}
  \end{subfigure}
  \hfill
  \begin{subfigure}[b]{0.12\linewidth}
    \centering
    \includegraphics[width=\textwidth]{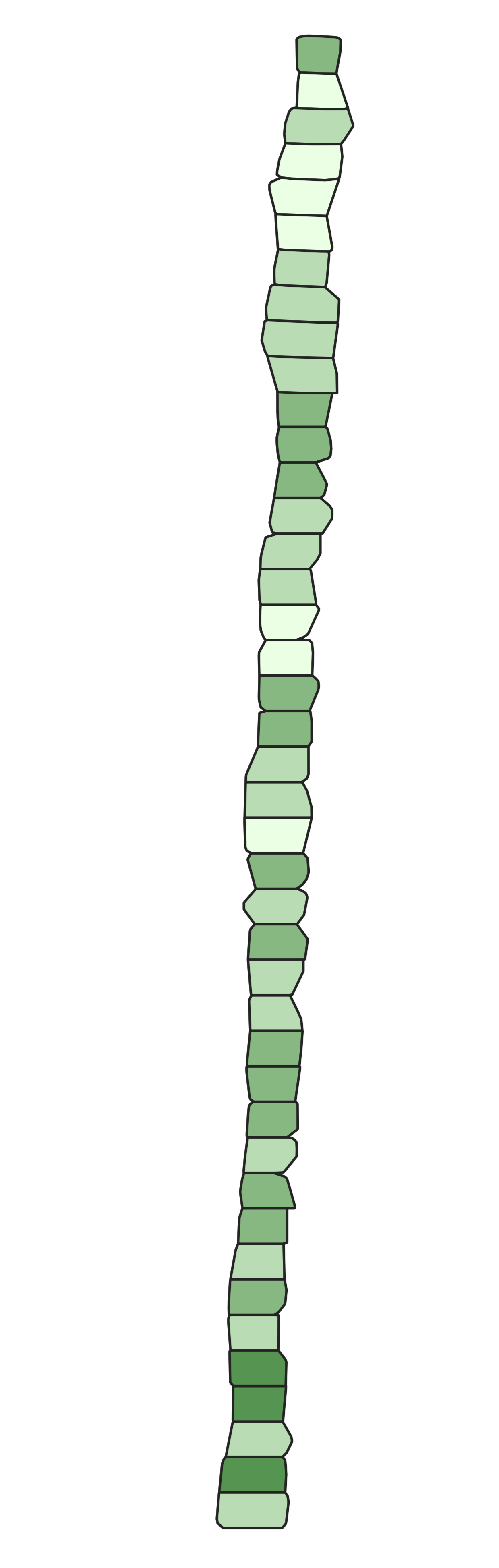}
    \caption{}
    \label{fig:20102}
  \end{subfigure}
  \hfill
  \begin{subfigure}[b]{0.08\linewidth}
    \centering
    \includegraphics[width=\textwidth]{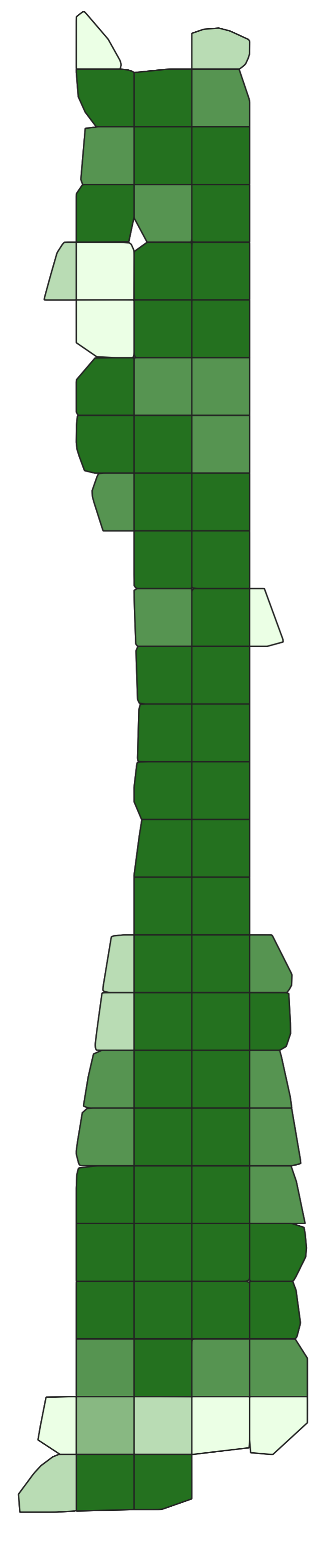}
    \caption{}
    \label{fig:k20}
  \end{subfigure}

\caption{Spatial distribution of the predicted AGB across plots based on real point cloud data derived from PointNet++ predictions trained on synthetic data downsampled with farthest point sampling: (a) MF1995.1, (b) MF1999.2, (c) MF2007.2, (d) HP2010.2, and (e) Knewleave20.}
\label{fig:predicted_agb}
\end{figure*}

Figure \ref{fig:predicted_agb} illustrates the spatial distribution of predicted AGB across five representative plots, derived from PointNet++ predictions trained on synthetic data downsampled with farthest point sampling, which achieved the highest accuracy among the tested models. The prediction range spans from 2.8 to 7.3 t, capturing variation in AGB across sites of different ages and sizes. The plots cover areas of 1.8 ha (MF1995.1), 1.0 ha (MF1999.2), 1.1 ha (MF2007.2), 0.8 ha (HP2010.2), and 2.5 ha (Knewleave20), representing diverse stand conditions. These examples demonstrate how PointNet++ effectively represents the spatial heterogeneity of forest structure.

\subsection{Comparison with other AGB estimation methods}
Indirect baselines include CHM-segmentation \citep{khoshelham2023tree}, three individual tree segmentation baselines (TreeLearn \citep{henrich2024treelearn}, SegmentAnyTree \citep{wielgosz2024segmentanytree}, and ForestFormer3D \citep{xiang2025forestformer3d}), and FullCAM \citep{stewart2023comparison}. Table \ref{tab:agb_carbon_comparison} reports plot-level AGB and carbon estimates for these methods alongside field measurements and the direct PointNet++ (FPS) results.

\begin{table*}[t]
\caption{Comparison of AGB and carbon (t/ha) estimates obtained by our direct method (PointNet++ with farthest point sampling datasets (FPS)), traditional methods, and field measurements for Jigsaw Farm (MF1995.1, MF1999.2, MF2007.2, HP2010.2) and Knewleave Farm (Knewleave20)}
\label{tab:agb_carbon_comparison}
\centering
\small 
\setlength\tabcolsep{3pt} 
\begin{tabular*}{\textwidth}{@{\extracolsep{\fill}} lcccccccccc @{}}
\toprule
\textbf{Method} & \multicolumn{2}{c}{\textbf{MF1995.1}} & \multicolumn{2}{c}{\textbf{MF1999.2}} & \multicolumn{2}{c}{\textbf{MF2007.2}} & \multicolumn{2}{c}{\textbf{HP2010.2}} & \multicolumn{2}{c}{\textbf{Knewleave20}} \\
\cmidrule(lr){2-3} \cmidrule(lr){4-5} \cmidrule(lr){6-7} \cmidrule(lr){8-9} \cmidrule(lr){10-11}
& AGB & C & AGB & C & AGB & C & AGB & C & AGB & C \\
& (t/ha) & (t/ha) & (t/ha) & (t/ha) & (t/ha) & (t/ha) & (t/ha) & (t/ha) & (t/ha) & (t/ha) \\
\midrule

CHM-segmentation & \multirow{2}{*}{79} & \multirow{2}{*}{40} & \multirow{2}{*}{70} & \multirow{2}{*}{35} & \multirow{2}{*}{50} & \multirow{2}{*}{25} & \multirow{2}{*}{55} & \multirow{2}{*}{28} & \multirow{2}{*}{67} & \multirow{2}{*}{34} \\[0.3ex]
\citep{khoshelham2023tree} & & & & & & & & & & \\
\addlinespace[1ex]

TreeLearn & \multirow{2}{*}{143} & \multirow{2}{*}{72} & \multirow{2}{*}{132} & \multirow{2}{*}{66} & \multirow{2}{*}{129} & \multirow{2}{*}{65} & \multirow{2}{*}{121} & \multirow{2}{*}{61} & \multirow{2}{*}{162} & \multirow{2}{*}{81} \\[0.3ex]
\citep{henrich2024treelearn} & & & & & & & & & & \\
\addlinespace[1ex]

SegmentAnyTree & \multirow{2}{*}{105} & \multirow{2}{*}{53} & \multirow{2}{*}{82} & \multirow{2}{*}{41} & \multirow{2}{*}{65} & \multirow{2}{*}{33} & \multirow{2}{*}{82} & \multirow{2}{*}{41} & \multirow{2}{*}{119} & \multirow{2}{*}{60} \\[0.3ex]
\citep{wielgosz2024segmentanytree} & & & & & & & & & & \\
\addlinespace[1ex]

ForestFormer3D & \multirow{2}{*}{89} & \multirow{2}{*}{45} & \multirow{2}{*}{55} & \multirow{2}{*}{28} & \multirow{2}{*}{43} & \multirow{2}{*}{22} & \multirow{2}{*}{48} & \multirow{2}{*}{24} & \multirow{2}{*}{69} & \multirow{2}{*}{35} \\[0.3ex]
\citep{xiang2025forestformer3d} & & & & & & & & & & \\
\addlinespace[1ex]

FullCAM & \multirow{2}{*}{---} & \multirow{2}{*}{33} & \multirow{2}{*}{---} & \multirow{2}{*}{31} & \multirow{2}{*}{---} & \multirow{2}{*}{25} & \multirow{2}{*}{---} & \multirow{2}{*}{18} & \multirow{2}{*}{---} & \multirow{2}{*}{23} \\[0.3ex]
\citep{stewart2023comparison} & & & & & & & & & & \\
\addlinespace[1ex]
\textbf{Our Direct method} & \multirow{2}{*}{\textbf{200} $\pm$ \textbf{2}} & \multirow{2}{*}{\textbf{100} $\pm$ \textbf{1}} & \multirow{2}{*}{\textbf{237} $\pm$ \textbf{2}} & \multirow{2}{*}{\textbf{119} $\pm$ \textbf{1}} & \multirow{2}{*}{\textbf{181} $\pm$ \textbf{1}} & \multirow{2}{*}{\textbf{91} $\pm$ \textbf{1}} & \multirow{2}{*}{\textbf{225} $\pm$ \textbf{4}} & \multirow{2}{*}{\textbf{113} $\pm$ \textbf{2}} & \multirow{2}{*}{\textbf{205} $\pm$ \textbf{7}} & \multirow{2}{*}{\textbf{103} $\pm$ \textbf{3}} \\[0.3ex]
\textbf{(PointNet++ (FPS))} & & & & & & & & & & \\
\addlinespace[1ex]
\cmidrule(lr){2-11}
Field measurement & 222 & 111 & 296 & 148 & 184 & 92 & 242 & 121 & 223 & 112 \\
\bottomrule
\end{tabular*}
\end{table*}

Across all sites, the indirect (allometry-based) baselines systematically underestimated plot-level carbon relative to field measurements, with the magnitude of underestimation varying by segmentation approach. The CHM-segmentation method underestimated carbon by 64\% to 77\%, while FullCAM showed even larger underestimation of 70\% to 85\% (e.g., 40 t/ha vs. 111 t/ha and 33 t/ha vs. 111 t/ha at MF1995.1, respectively). Among deep-learning individual tree segmentation baselines, TreeLearn produced the smallest discrepancies (27\% to 55\% below), followed by SegmentAnyTree (47\% to 72\% below) and ForestFormer3D (60\% to 81\% below). In contrast, our direct PointNet++ (FPS) estimates were consistently closest to the field measurements, with discrepancies of only 2\% to 20\% below across the five plots, indicating substantially improved agreement compared to all indirect baselines.
These results demonstrate the significant advantages of deep learning approaches in capturing complex spatial and structural details from point cloud data, resulting in more accurate and reliable carbon stock estimates than traditional indirect methods like individual tree segmentation baselines and FullCAM.

\section{Discussion}\label{sec5}
\subsection{Dataset characteristics and downsampling effect}
As shown in Table \ref{tab:downsampling_comparison}, both random sampling and farthest point sampling produced synthetic datasets with identical point counts (2048 points per plot) but differing spatial distributions. Farthest point sampling ensured broader spatial coverage and produced a more uniform spatial distribution than random sampling (Table \ref{tab:downsampling_comparison}). These differences in spatial representation had a clear impact on volume estimation for the real dataset, as illustrated in Figure \ref{fig:fpsvsfxreal}.
While the two downsampling methods performed comparably on synthetic data (Figure \ref{fig:fpsvsfxsynth}), farthest point sampling consistently generalized better to real point clouds by reducing the underestimation biases observed with random sampling. In our experiments, farthest point sampling produced higher AGB estimates than random sampling and was generally closer to field measurements (Tables \ref{tab:fps_agbc} and \ref{tab:rs_agbc}), demonstrating its advantage in preserving structural integrity and improving estimation accuracy. The underlying reason for this improvement lies in the spatial distribution of sampled points. With farthest point sampling, the average neighboring distance between points was approximately 50 to 60 percent greater than with random sampling (Table \ref{tab:downsampling_comparison}). This wider spacing enabled the models to capture the three-dimensional structure of trees more effectively, as the sparser points encouraged them to consider broader spatial extents and reduced the compressed representation of tree shape. In contrast, random sampling generated denser local clusters, which often caused the models to interpret trees as more compact by overfitting to local details and missing outer boundaries, thereby underestimating tree volumes.

These findings highlight the critical role of downsampling strategy in enabling deep learning models to adapt to real point cloud data, with farthest point sampling emerging as the more effective method. Our results are also consistent with \citet{qi2017pointnet++}, who demonstrated that farthest point sampling adapts effectively to diverse input distributions, producing receptive fields that capture geometric features at multiple scales. Such characteristics are particularly important for downstream tasks such as segmentation, classification, and regression, which require strong generalization across varying data distributions.

\subsection{Interpretation of synthetic data performance}
The evaluation of the deep regression models on synthetic datasets showed the remarkable performance of PointNet++, DGCNN, and PointConv, which achieved an  MAPE of approximately 2\% on synthetic validation plots. 
PointNet achieved higher validation error, consistent with its limited ability to encode local neighbourhood structure.
The superior performance of PointNet++, DGCNN, and PointConv can be attributed to their ability to capture local neighborhood information within point cloud data, a critical feature absent in the original PointNet architecture. PointNet treats each point independently, without directly encoding spatial relationships, which limits its ability to capture fine-grained structural details. By contrast, PointNet++, DGCNN, and PointConv incorporate neighborhood information, enabling better preservation and understanding of spatial structures.

\subsection{Real data transfer and baseline comparison}

The deep regression models significantly outperformed traditional indirect methods. Among the direct models, PointNet++ (FPS) was closest to the field measurements, with discrepancies of approximately 2\%--20\% across the five plots. Among the indirect baselines, TreeLearn, a deep-learning individual tree segmentation baselines, showed the most competitive performance; however, it still exhibited substantial underestimation of AGB and carbon, with discrepancies ranging from 27\% to 55\% compared to field measurements. Other traditional approaches fared worse: the CHM-segmentation method consistently underestimated these metrics by 64\% to 77\%, while FullCAM showed carbon discrepancies between 70\% and 85\%. It should be noted that the field-based reference values were derived from standard allometric equations; although widely used, this ground-truthing approach is not necessarily unbiased and can propagate allometric-model uncertainty into the reported reference AGB/carbon values.

While individual tree segmentation baselines can achieve high accuracy when trees are well separated and segmentation is reliable, their performance can degrade under crown overlap, occlusion, or platform/sensor changes because errors propagate from segmentation to attribute extraction and then to allometric conversion \citep{ma2021novel, zhang2024individual}. Our plot-level regression avoids explicit instance delineation and can therefore be applied in settings where individual tree segmentation is unstable or costly to validate, while still producing plot-level AGB and carbon estimates.
These findings are consistent with \cite{ma2025framework}, who highlighted that individual tree segmentation baselines AGB estimation depends on several sequential components and associated uncertainties (e.g., delineation, attribute estimation, and allometric models), which can accumulate and propagate when upscaling from trees to plots and larger areas. This limitation reinforces the advantages of deep learning approaches, which offer scalability and adaptability for large-scale ecological studies \citep{paul2016testing}.
A key strength of the methodology employed in this study lies in the creation of realistic synthetic datasets to train deep learning models, which are then applied to real point cloud data. Synthetic data enables the generation of diverse, large-scale datasets that capture a wide range of forest structures and conditions, addressing the challenges posed by limited availability of field measurements. By training models on synthetic data, deep learning architectures such as PointNet++, PointConv, and DGCNN can learn critical features related to AGB and carbon estimation in a controlled and scalable environment. This approach reduces dependency on extensive field data collection, which is often labor-intensive and time-consuming, while ensuring the models generalize effectively to real-world scenarios.

Moreover, the synthetic datasets used in this study were designed to closely mimic the characteristics of real forest plots (Table \ref{tab:downsampling_comparison}), allowing the trained models to perform with high confidence when applied to real data. This strategy ensures that the models are not only effective but also scalable for large-scale forest AGB and carbon stock assessments. The integration of synthetic data significantly enhances the efficiency and practicality of machine learning in ecological studies, offering a robust solution for regions where field-based datasets are sparse or inaccessible. This approach demonstrates the transformative potential of synthetic data in addressing data scarcity, making it a valuable tool for advancing large-scale ecological and environmental research.

\subsection{Limitations and future work}
Despite the promising results, this study has several limitations. The deep learning models were trained exclusively on synthetic data, which may not fully capture the complexity and variability of real-world forest plots. Differences in point density, noise, and occlusion between simulated and real scans may introduce domain shift and reduce transferability to new sites. Our synthetic plots were designed to mimic the real sites in terms of acquisition modality (ULS), approximate point density, and forest type (eucalyptus-dominated stands). Nevertheless, synthetic scenes cannot fully reproduce real-world variability such as sensor-specific noise, multipath/beam divergence effects, imperfect ground normalization, and complex occlusion patterns in multilayer canopies, which can lead to distribution shift at inference time. We did not apply explicit domain adaptation between synthetic and real LiDAR distributions in this study; this is a priority for future work to further reduce residual bias on real sites. Additionally, this study focused on woody components and excluded leaves, which may lead to systematic underestimation of AGB and carbon.

Model performance was also influenced by the choice of downsampling method. While farthest point sampling provided better generalization, random sampling tended to over-represent locally dense regions, which can bias predicted volumes when transferring to real data. Moreover, the higher variability observed in PointNet indicates that not all point-based architectures are equally robust for plot-level AGB estimation.

To address these limitations, future work should reduce synthetic–real domain shift by incorporating more heterogeneous real point clouds (even a small calibration set) and by explicitly simulating realistic noise, occlusion, and density patterns during synthetic data generation. Extending the synthetic scenes to include foliage, additional species, and broader structural variability (e.g., age classes and stocking) would further improve realism. Finally, future studies should evaluate domain-adaptation or lightweight calibration (e.g., site-level scaling) and report uncertainty propagation from volume prediction to AGB/carbon conversion to improve reliability in operational settings.

\section{Conclusion}
This study demonstrated the effectiveness of integrating synthetic point cloud data with deep learning models to directly estimate wood volume, AGB, and carbon stocks. By leveraging synthetic data for model training, we successfully applied deep learning architectures to real point cloud datasets, achieving significantly higher accuracy compared to traditional indirect methods. 
Our results highlight a clear performance gap: the process-based FullCAM tool underestimated carbon stocks by 70\% to 85\%, while the CHM-segmentation method exhibited discrepancies of 64\% to 77\%. Among the deep-learning individual tree segmentation baselines, TreeLearn was the most accurate, yet still underestimated carbon stocks by 27\% to 55\%. In contrast, PointNet++ (FPS) provided the strongest precision and generalization, with discrepancies relative to field measurements limited to 2\% to 20\%.
These findings mark a significant advancement in forest carbon stock estimation, with synthetic data enabling the creation of diverse, scalable training datasets that support model generalization despite challenges associated with domain shifts. 
Our results also demonstrate the importance of the downsampling method in preserving the spatial integrity of both synthetic and real datasets. farthest point sampling consistently yielded results closer to field measurements, providing broader spatial coverage and capturing more diverse sections of the dataset, making it the preferred method for accurate and reliable AGB and carbon stock predictions.
Furthermore, this study underscores the potential of deep learning models to reduce dependence on labor-intensive field measurements while maintaining high accuracy and consistency in AGB and carbon stock predictions. This novel approach of combining synthetic data generation with deep learning provides an efficient and scalable solution for forest monitoring, offering substantial improvements over traditional methods.
However, challenges remain, particularly in addressing domain adaptation and incorporating non-woody AGB components into the models. Future research should prioritize refining model performance by incorporating more heterogeneous datasets to capture a broader range of forest structures and conditions. Additionally, strategies to mitigate domain shifts between synthetic and real data will be critical for enhancing model accuracy and reliability.
Overall, this study highlights the transformative potential of synthetic data and deep learning for large-scale forest AGB estimation. By further improving these methods and integrating optimal downsampling strategies, they can become invaluable tools for efficient, accurate, and accessible forest monitoring on a global scale.

\section*{Acknowledgments}
This project was partially funded by the Australian Government’s Department of Agriculture, Fisheries \& Forestry (DAFF). Project lead, The University of Melbourne, is a partner in the Victoria Drought Resilience Adoption \& Innovation Hub, which is supported by DAFF. The first two authors acknowledge the financial support from the University of Melbourne through the Melbourne Research Scholarship. This research is supported by The University of Melbourne’s Research Computing Services. Special thanks to Zexian Huang for his significant contributions to the coding and computational analysis, which were essential to the success of this research.

{
	\begin{spacing}{1.17}
		\normalsize
		\bibliography{ISPRSguidelines_authors} 
	\end{spacing}
}

\end{document}